\documentclass{article}

    \PassOptionsToPackage{numbers, compress}{natbib}
    \usepackage[final]{neurips_2020}

\usepackage[utf8]{inputenc} % allow utf-8 input
\usepackage[T1]{fontenc}    % use 8-bit T1 fonts
\usepackage{hyperref}       % hyperlinks
\usepackage{url}            % simple URL typesetting
\usepackage{booktabs}       % professional-quality tables
\usepackage{amsfonts}       % blackboard math symbols
\usepackage{nicefrac}       % compact symbols for 1/2, etc.
\usepackage{microtype}      % microtypography
\usepackage{natbib}

\usepackage{color}
\usepackage{amsmath}
\usepackage{amssymb}
\usepackage{graphicx}
\usepackage[linesnumbered, ruled]{algorithm2e}
\definecolor{ballblue}{rgb}{0.13, 0.67, 0.8}
\definecolor{bluegray}{rgb}{0.4, 0.6, 0.8}

\SetCommentSty{mycommfont}
\usepackage{multirow}
\usepackage{tabularx}
\usepackage{adjustbox}

\newcommand{\figref}[1]{Fig.~\ref{#1}}

\newcommand{\tabref}[1]{Tab.~\ref{#1}}
\newcommand{\eqnref}[1]{Eq.~(\ref{#1})}
\newcommand{\secref}[1]{Sec.~\ref{#1}}
\newcommand{\adxref}[1]{Appendix \ref{#1}}

\title{An Efficient Asynchronous Method for Integrating Evolutionary and Gradient-based Policy Search}

\author{%
  Kyunghyun Lee \qquad Byeong-Uk Lee \qquad Ukcheol Shin \qquad In So Kweon \\
  Korea Advanced Institute of Science and Technology (KAIST)\\
  Daejeon, Korea \\
  \texttt{\{kyunghyun.lee, byeonguk.lee, shinwc159, iskweon77\}@kaist.ac.kr}
}

\begin{document}

\maketitle

\begin{abstract}
%Deep reinforcement learning (DRL) algorithms and evolution strategies (ES) have been applied to various tasks, showing excellent performances.
%These have the opposite properties, with DRL having good sample efficiency and poor stability, while ES being vice versa.
%Recently, there have been attempts to combine these algorithms not regarding them as competitors but taking advantage of each other.
%However, these previous combined algorithms fully rely on synchronous update scheme that could cause a meaningless idle time of each thread and low exploration property.
%To solve this challenge, asynchronous update scheme is one of the possible solution; time-efficient and high exploration property, especially when each agent has diverse episode lengths.
%Further, this asynchronous update scheme encourages more diverse policy exploration as it enables immediate updating of the population
%In this paper, we introduce Asynchronous Evolution Strategy-Reinforcement Learning (AES-RL) that maximizes the parallel efficiency of ES and integrated it with policy gradient methods.
%Specifically, we propose 1) a novel framework to merge ES and DRL asynchronously and 2) a few asynchronous update methods that can take all benefits of asynchronism, ES, and DRL: exploration, stability, and sample efficiency, respectively.
%The proposed framework and update methods are evaluated in continuous control benchmark work, showing superior performance as well as time efficiency compared to the previous methods.
Deep reinforcement learning (DRL) algorithms and evolution strategies (ES) have been applied to various tasks, showing excellent performances.
These have the opposite properties, with DRL having good sample efficiency and poor stability, while ES being vice versa.
Recently, there have been attempts to combine these algorithms, but these methods fully rely on synchronous update scheme, making it not ideal to maximize the benefits of the parallelism in ES.
To solve this challenge, asynchronous update scheme was introduced, which is capable of good time-efficiency and diverse policy exploration.
In this paper, we introduce an Asynchronous Evolution Strategy-Reinforcement Learning (AES-RL) that maximizes the parallel efficiency of ES and integrates it with policy gradient methods.
Specifically, we propose 1) a novel framework to merge ES and DRL asynchronously and 2) various asynchronous update methods that can take all advantages of asynchronism, ES, and DRL, which are exploration and time efficiency, stability, and sample efficiency, respectively.
The proposed framework and update methods are evaluated in continuous control benchmark work, showing superior performance as well as time efficiency compared to the previous methods.
\end{abstract}

\section{Introduction}
Reinforcement Learning (RL) algorithms, one major branch in policy search algorithm, were combined with deep learning and showed excellent performance in various environments, such as playing simple video games with superhuman performance~\cite{ataridqn13, a3c16}, mastering the Go~\cite{alphazero17}, and solving continuous control tasks~\cite{ddpg2015, td32018, haarnoja2018soft}.
Evolutionary methods, another famous policy search algorithm, were applied to the parameters of deep neural network and showed compatible results as Deep Reinforcement Learning (DRL)~\cite{es_salimans2017}.

These two branches of policy search algorithms have different properties in terms of sample efficiency and stability~\cite{policy_search2018}.
%While DRL is sample efficient by learning from every step of an episode, Evolution Strategies (ES) are the opposite because they learn from the results of the whole episode~\cite{policy_search2018}.
%In particular, the off-policy DRL algorithm with replay memory learns several times from a single step by using a replay buffer.
%However, off-policy DRL algorithms like Deep Deterministic Policy Gradient (DDPG) are more sensitive to hyperparameters and often diverge~\cite{convergent2009, reproduce2017, stability2019, DBLP:journals/corr/abs-1709-06560}.
DRL is sample efficient, since it learns from every step of an episode, but is sensitive to hyperparameters~\cite{convergent2009, reproduce2017, stability2019, DBLP:journals/corr/abs-1709-06560}.
Evolution Strategies (ES) are often considered as the opposite because they are relatively stable, learning from the result of the whole episode~\cite{policy_search2018}, yet they require much more steps in the learning process~\cite{policy_search2018, erl2018, pderl2019}.
% Although there is an attempt to reuse samples with ``importance mixing" mechanism in ES for better efficiency, it is still far behind DRL~\cite{importance2018}. 

ES and DRL are often considered as competitive approaches in policy search~\cite{es_salimans2017, neuro_ga_2017}, and relatively few studies have tried to combine them~\cite{houthooft2018evolved, Ackley+Littman:1992, DRUGAN2019228}.
% \KH{too offensive. only a few -> relatively.}
Recently, some works tried to utilize the useful gradient information of DRL into ES directly.
Evolutionary Reinforcement Learning (ERL) has an independent RL agent that is periodically injected into the population~\cite{erl2018}.
In Cross-Entropy Method-Reinforcement Learning (CEM-RL), half of the population is trained with gradient information, and new mean and variance of the population are calculated with better performing individuals~\cite{pourchot2018cemrl}. 

%In both ES and RL, synchronous parallelization methods were introduced for faster learning and stability~\cite{a3c16, es_salimans2017, neuro_ga_2017, espeholt2018impala, apex2018}.
% \KH{both works -> ES and RL. looks like ERL and CEM-RL}
%However, a common problem with parallelization is that when the length of episodes from multiple agents differ, the calculation time is aligned to an agent with the longest duration.
%In particular, the synchronous ES methods, including previously suggested ERL approaches, suffer from this time inefficiency because they generate new populations only after evaluation of all agents are done.
In both ES and RL, parallelization methods were introduced for faster learning and stability~\cite{a3c16, es_salimans2017, neuro_ga_2017, espeholt2018impala, apex2018}.
Most of these parallelization methods take synchronous update scheme, which aligns the update schedule of every agents to the one with the longest evaluation time.
This causes crucial time inefficiency because agents with shorter evaluation should wait until the whole agents finish their job.

The asynchronous method is one of the direct solutions to this problem, as one agent can start the next evaluation immediately without waiting other agents~\cite{aga1993, pnsga2008, Martin2015AsynchronousPE, Scott2015UnderstandingSA}.
Another advantage of the asynchronous method is that updates occur more often than those of synchronous methods, which can encourage diverse exploration~\cite{Scott2015UnderstandingSA, ANES2013}.

In this paper, we propose a novel asynchronous framework that efficiently combines both ES and DRL, alongside with some effective asynchronous update schemes by thoroughly analyzing the property of each.
The proposed framework and update schemes are evaluated on the continuous control benchmark, underlining its superior performance and time-efficiency compared to previous ERL approaches.

Our contributions include the following:
\begin{itemize}
\item
We propose a novel asynchronous framework that efficiently combines both Evolution Strategies (ES) and Deep Reinforcement Learning algorithms. 
\item
We introduce several asynchronous update methods for the population distribution. We thoroughly analyze all update methods and their properties.
Finally, we propose the most effective asynchronous update rule. 
\item
We demonstrate the time and sample efficiency of the proposed asynchronous method.
The proposed method reduces the entire training time about 75\% wall clock on the same hardware configuration.
Also, the proposed method can achieve up to 20\% score gain with given time steps through effective asynchronous policy searching algorithm. 
\end{itemize}

\section{Background and Related work}
%\subsection{Evolutionary Algorithms}
\subsection{Evolution Strategies}
\label{sec:EA}
%Evolutionary Algorithm (EA) is a type of black-box optimization inspired by natural evolution that consists of three major evolutionary operators: \textit{selection}, \textit{recombination}, and \textit{mutation}.
%The \textit{selection} operator choose candidate for next generation.
%New offspring are created from selected parents by the \textit{recombination} operator.
%The \textit{mutation} operator generates random change, like Gaussian noise.
%All individuals are evaluated to calculate a fitness value, which is the value we want to optimize.
%The fitness function $f: \mathbb{R}^d \rightarrow \mathbb{R}$ maps parameters to fitness value.
%The objective of EA is similar to that of RL, where the reward in the RL is equal to the fitness value of the EA.
Evolutionary Algorithm (EA) is a type of black-box optimization inspired by natural evolution, that aims to optimize a fitness value.
All individuals in EA are evaluated to calculate fitness function $f$, which is similar concept to the reward in RL.
Evolutionary Strategy (ES) is one of the main branches in EA that one individual remains for each generation.
When the population is represented by mean and covariance matrix, these algorithms are called estimation of distribution.
The most famous algorithms in this category are the Cross-Entropy Method (CEM)~\cite{cem2004} and Covariance Matrix Adaptation Evolution Strategy (CMA-ES)~\cite{cmaes2016}. 

In CEM, individuals are sampled based on the distribution $\mathcal{N}(\mu, \mathbf{\Sigma})$.
All individuals are evaluated and the distribution is updated based on the fixed number of best-performing individuals.
CMA-ES is similar to CEM, except it considers the ``evolutionary path'' that collects the direction of consecutive generations.

\subsection{Parallel, Synchronous and Asynchronous ES}
%Parallelization is the most useful way to increase the learning speed of the ES and DRL.
%In general, most of the time is consumed in the evaluation stage of each individual.
%Parallelization can reduce time with concurrent execution.
Parallelization is the most useful way to increase the learning speed of the ES and DRL, by enabling concurrent execution.
%Changing from serial to synchronous parallel has no effects on algorithmic flow but can expect substantial improvement in execution speed.
%\citet{es_salimans2017} solved MuJoCo humanoid with a simple ES algorithm in 10 minutes by using 1440 cores.
%Also,~\citet{neuro_ga_2017} used 720 cores for a genetic algorithm.
%Many existing studies reported improvements using parallelization~\cite{pnsga2008, ANES2013, asurvey1998, surveypal1999, pgatheory2013}.
Many existing studies reported improvements using parallelization~\cite{pnsga2008, ANES2013, asurvey1998, surveypal1999, pgatheory2013}.
For example, \citet{es_salimans2017} solved MuJoCo humanoid with a simple ES algorithm in 10 minutes by using 1440 parallel cores.
%However, it is hard to expect that computation speed will increase linearly as the number of cores increases~\cite{asummary1995}, especially in a heterogeneous environment where the evaluation time varies. 
%This phenomenon is caused by synchronization process in many parallel EAs.
In particular, changing from serial to synchronous parallel requires no modification on algorithmic flow, while enabling substantial improvement in execution speed.

The synchronization process is a step that updates the population with the results of all individuals for the next generation.
Therefore, when the evaluation times of all individuals differ, all workers should wait until the last worker is finished~\cite{Scott2015UnderstandingSA}, thereby increasing idle time.
This is more critical when the number of cores increases, since it is hard to expect linear increase of time efficiency~\cite{Martin2015AsynchronousPE}.

\begin{figure}[t]
  \centering
  \includegraphics[width=\textwidth]{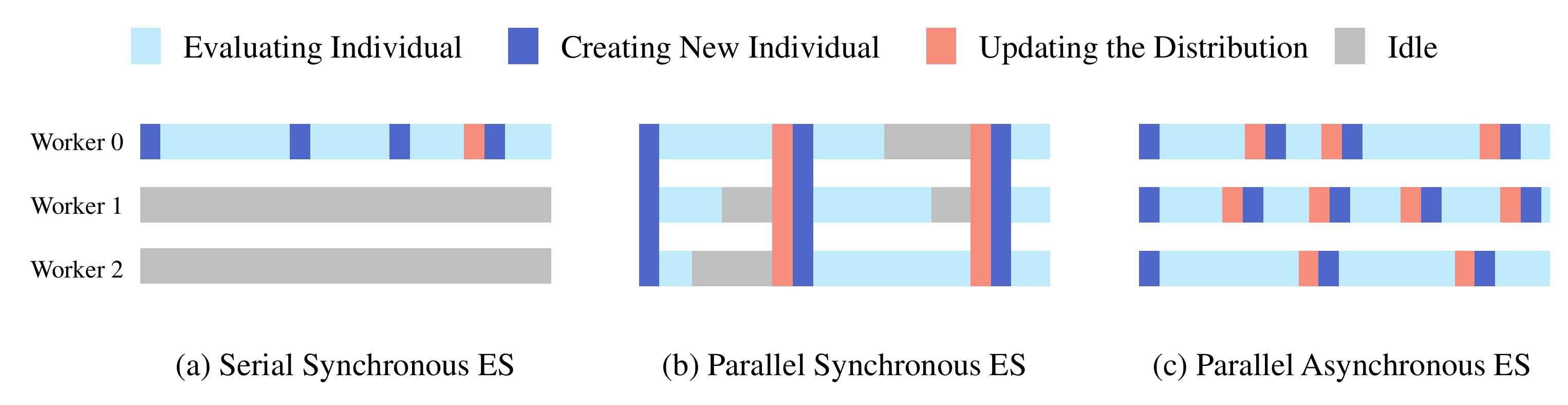}
  \vspace{-0.25in}
  \caption{{\bf Parameter Update Scheme Comparison.} (a) One worker is used in serial synchronous ES. The worker evaluate the individuals sequentially. (b) In parallel ES, all workers evaluate the individuals in parallel. However, some workers are in idle until the last worker is finished in each generation. (c) All workers update the distribution immediately after the evaluation and evaluate the next generation. Note that the number of updates per evaluation is much more frequent in (c) compare to the other methods, which encourages the exploration. }
  \label{fig:overview}
  \vspace{-0.2in}
\end{figure}

Asynchronous algorithms are introduced to address this problem. 
%The algorithms eliminate synchronization processes. 
%Instead, they propose a modified process to allow all individuals to update the population immediately after the evaluation without waiting other workers, letting all workers to evaluate a parameter without idle time.
They propose a modified process to allow all individuals to update the population immediately after their evaluation, thereby reducing meaningless idle time.
Also, more frequent updates increase the exploration in the parameter space~\cite{Scott2015UnderstandingSA, Martin2015AsynchronousPE}.
Concepts and differences are illustrated in \figref{fig:overview}.
% An ES algorithm limited the maximum episode length based on the average length of the population, achieving the average usage of cores more than 50\%~\cite{es_salimans2017}.

\subsection{Update methods in ES} \label{sec:updates}
The update method can be divided into two main approaches depending on how the fitness value is used: the rank-based methods and the fitness-based methods.
The rank-based methods sort the fitness values of all individuals and only the rank information is used for update.
The fitness-based methods use the fitness value itself.
The rank-based methods are more conservative and robust to inferiors.
However, it also ignores the useful information from superiors~\cite{selection1995}. 

In synchronous and asynchronous perspective, rank-based methods are more suitable to synchronous methods because it needs the results of all individuals in the population.
%There is an asynchronous version of the rank-based update by preserving past individuals~\cite{ANES2013}.
On the other hand, fitness-based methods do not have this restriction and can utilize the superior individual sufficiently. % as opposite to the rank-based methods.
The main problem in fitness-based methods is fitness value normalizing, because the evolution pressure is directly proportional to the fitness value.

In CEM, the population is updated with the rank-based method, as
\begin{equation}
\label{eq:covarcalc}
\mu_{t+1} = \sum_{i=1}^{K_e}\lambda_i z_i, \qquad \mathbf{\Sigma}_{t+1} = \sum_{i=1}^{K_e}\lambda_i (z_i -\mu_t)(z_i - \mu_t)^{\intercal} + \epsilon \mathcal{I},
\end{equation}
where $K_e$ is a number of elite individuals, $(z_i)_{i=1,...,K_e}$ are parameters of selected individuals according to the fitness value, and $(\lambda_i)_{i=1,...,K_e}$ are weights of selection for the individuals.
These weights are generally either $\lambda_i = \frac{1}{K_e}$ or $\frac{\log(1+K_e)/i}{\sum_{i=1}^{K_e}\log(1+K_e)/i}$~\cite{cmaes2016}.
The former is the method of giving the same weights for the selected individuals, and the latter is the method of giving higher weights according to rank.

The covariance matrix calculation is sometimes reduced to variance because the calculation complexity grows exponentially as the number of parameters increases~\cite{pourchot2018cemrl}.
Then, \eqnref{eq:covarcalc} is simplified as
\begin{equation}
\mathbf{\Sigma}_{t+1} = \sum_{i=1}^{K_e}\lambda_i (z_i -\mu_t)^2 + \epsilon \mathcal{I}.
\label{eq:simplecovalcalc}
\end{equation}

% These weights are generally chosen as the same weights for the selected individuals, or as higher weights according to rank~\cite{cmaes2016}.

%\textcolor{red}{Rank based and fitness based methods}

%\subsection{DDPG and Twin-Delayed DDPG (TD3)}
%DDPG~\cite{ddpg2015} and TD3~\cite{td32018} are actor-critic off-policy algorithms and known to have good sample efficiency. However, DDPG is unstable because of the bias from overestimating in critic updates~\cite{convergent2009, DBLP:journals/corr/abs-1709-06560}. TD3 improves DDPG with an additional critic, and prevent overestimation by taking lower estimates from two critics. 

\subsection{Evolutionary Reinforcement Learnings}
%Evolutionary algorithm (EA) and deep reinforcement learning (DRL) algorithms are two popular approaches for the policy search. 
As mentioned in the introduction section, EA and DRL have opposite property in sample efficiency and stability.
%EA shows low sample efficiency but is robust to hyperparameter change since it learns from the whole episode.
%On the other hand, DRL has high sample efficiency yet is highly sensitive to hyperparameter due to learning from each step of an episode. 
Based on these characteristics, there have been a few approaches to combine these two, taking advantages from both~\cite{erl2018,khadka2019collaborative,pderl2019,pourchot2018cemrl}.

The evolutionary reinforcement learning (ERL)~\cite{erl2018} is the pioneering trial to merge two approaches: population-based evolutionary algorithm and sample efficient off-policy deep RL algorithm (DDPG)~\cite{ddpg2015}.
%Based on these combinations, ERL was evaluated on robot locomotion tasks and showed the practical benefits of merging the two approaches. 
However, the sample efficiency issue remains an important problem, which triggered various enhanced version of ERL to be suggested.
Collaborative Evolutionary Reinforcement Learning (CERL)~\cite{khadka2019collaborative} collects different time-horizons episodes by leveraging a portfolio of multiple learners to improve sample diversity and sample efficiency.
Further, Proximal Distilled Evolutionary Reinforcement Learning (PDERL)~\cite{pderl2019} points out that the standard operators of GA, the base evolutionary algorithm of ERL, are destructive and cause catastrophic forgetting of the traits.
%/UK{PDERL proposed a new hierarchical integration learning method between evolution and learning along with learning-based variation operators for compensating the problem of the genetic representation.}
%\BU{PDERL proposed a new hierarchical integration of evolutionary learning with learning-based variation of operators for compensating the problem of the genetic representation.}
%PDERL proposed a new hierarchical learning method and learning-based variation operators.
All ERL variants share the same architecture, an independent RL agent.
That is, an RL agent is trained along with the ES, and injected into the population periodically to leverage the RL. 
Otherwise, CEM-RL~\cite{pourchot2018cemrl} trains the half of individuals with RL algorithm, utilizing the gradient information directly.

% CEM-RL~\cite{pourchot2018cemrl} is a joint framework that exploits the Cross-Entropy Method (CEM) and Twin Delayed Deep Deterministic Policy Gradient (TD3~\cite{td32018}).
% CEM-RL focuses on recovering the sample efficient property of the EA algorithm itself, such as search efficiency and importance mixing of ES by utilizing the cross-entropy method and improved version of off-policy update method.
% \todo{Modify this paragraph about CEM-RL}
\section{Methods}
\label{sec:methods}

%We combined various update methods in ES with RL in an asynchronous way. 
%Our algorithm is composed of ES and RL parts. 
%Any kind of the off-policy actor-critic algorithm structure with a replay buffer can be applied to the RL part.
%In this paper, we used TD3 in all experiments, because the choice of a RL algorithm is not the main contribution. \UK{to experiment part}
%For ES parts, we use various update methods, some based on the previous works, and others are newly conceived. 
In this section, we present a novel asynchronous merged framework of ES and RL. 
Any kind of the off-policy actor-critic algorithm with a replay buffer can be applied to the RL part.
Also, we present effective new update method that balances between exploration and stability.
A pseudocode for the whole algorithm is described in Appendix \ref{apx:algorithm}.

%\UK{In this section, we present a novel asynchronous merged framework of ES and RL. Any kind of the off-policy actor-critic algorithm with a replay buffer can be applied to the RL part. Also, we present effective asynchronous update rules that can collect valuable samples.}%including few update rules used only in the previous ES literature.}

\subsection{Asynchronous Framework}
\begin{figure}[t]
  \centering
  \includegraphics[width=\textwidth]{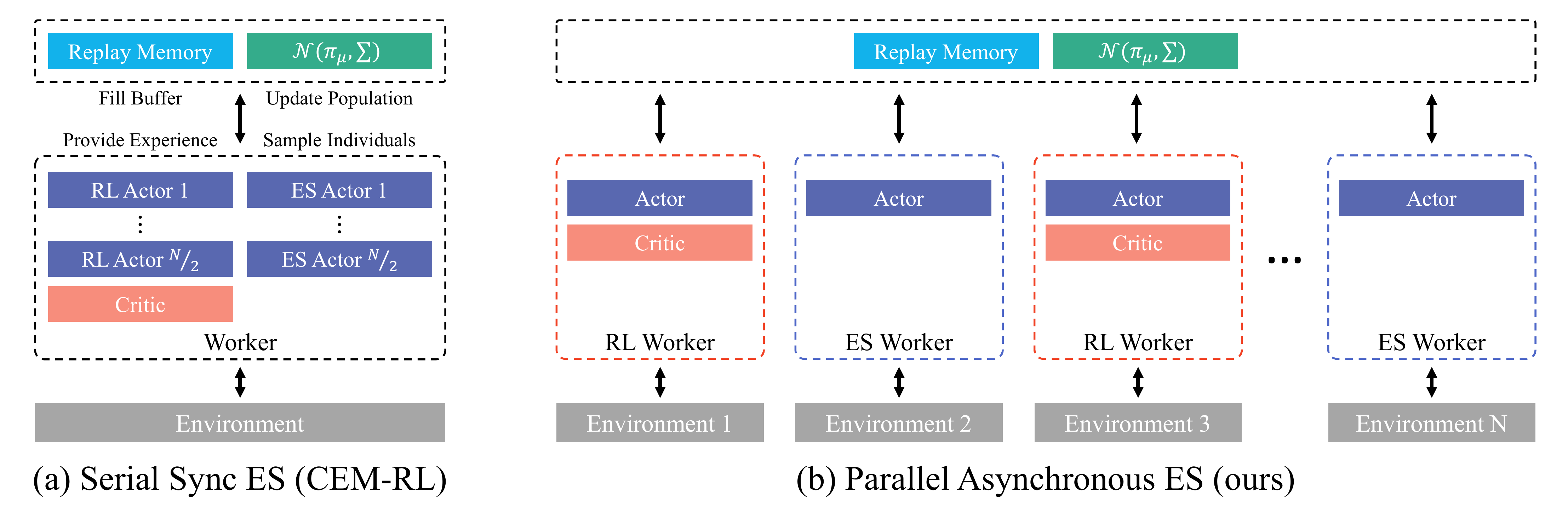}
  \vspace{-0.25in}
  \caption{{\bf Framework Comparison with Serial Synchronous (CEM-RL) and Parallel Asynchronous method (ours).} CEM-RL has one worker and evaluates all individuals sequentially. Ours has several distributed workers and evaluates individuals asynchronously}
  \label{fig:async_framework}
  \vspace{-0.2in}
\end{figure}

As described in~\figref{fig:overview}, the parallel asynchronous update scheme is more time-efficient and encourages exploration compared to the synchronous update schemes. 
The framework of our baseline, CEM-RL, has half ES actors and half RL actors as shown in~\figref{fig:async_framework}-(a).
Only after all evaluations of the actors are finished, CEM-RL updates population distribution represented by the mean $\mu$ and the variance $\mathbf{\Sigma}$.
On the other hand, our proposed parallel asynchronous framework allows each worker to have an ES or RL actor and individually update the population distribution as shown in~\figref{fig:async_framework}-(b).
Also, in CEM-RL, the shared critic network learns after synchronizing step and takes time for a while.
In contrast, the shared critic network continuously learns in parallel with the other actors in our framework.
Based on the change of this update scheme, the overall training time reduction and active exploration can be expected.

However, there are some problems to convert the serial synchronous update scheme to the parallel asynchronous update scheme.
The first problem occurs in creating new individuals.
CEM-RL creates a fixed number of ES and RL actors in new generation only after all actors are terminated.
It is hard to apply the method directly to the asynchronous update because the actors have different ending moments. 
% But in the case of the asynchronous update, it is impossible to use this method because the actors have different ending moments. 
This problem is handled in \secref{sec:popcon}.
Secondly, we need an asynchronous method to update population distribution effectively. 
Since each actor has its own fitness value, a novel update method to adaptively exploit this is required. 
This problem is handled in \secref{sec:MU} and \secref{sec:VMU}.
Our overall algorithm pseudo-code is prested in \adxref{apx:algorithm}

\subsection{RL and ES Population Control}
\label{sec:popcon}
%We introduce the probability based population control method for asynchronous algorithms. 
%We define a $p_\text{rl}$ that is the probability of being an RL actor. 
%It is adjusted according to the past number of RL and ES actors.
%We used a simple $P$ controller to make the RL agent ratio converge to the desired rate.
%As a result, $p_{\text{rl}}$ can be described as,
To solve the new individual creating problem, we introduce a probability based population control method for the asynchronous scheme.
This method probabilistically allocates a new individual as an ES or RL actor so that the ratio between RL and ES actor is maintained as a desired ratio.
The probability of being an RL actor is represented as $p_{\text{rl}}$, and is adaptively determined by the cumulative number of RL and ES actors, described as follows:
%\UK{To solve the new individuals creating problem, we introduce the probability based population control method for the asynchronous scheme.
%This method allocates a new individual as an ES or RL actor so that the ratio between RL and ES actor is maintained as a desired ratio.
%The ES or RL actor is determined according to the probability $p_{\text{rl}}$, the probability of being an RL actor. The probability $p_{\text{rl}}$ %is adaptively determined by the current number of RL and ES actors, and described as follows:
%}
\begin{equation}
p_{\text{rl}} = \text{clip}\left(-K_{rl} \left[\frac{n_{\text{rl}}}{n_{\text{rl}} + n_{\text{es}}}  - p_{\text{desired}}\right] + 0.5, 0, 1\right)
\label{eq:prl}
\end{equation}
where $K_{\text{rl}}$ is a $P$ controller gain, $p_\text{desired}$ is the desired rate of RL agent, and $n_{\text{rl}}$ and $n_{\text{es}}$ are the total number of RL and ES agents, respectively.
%\UK{Need to state default value that used in all experiments or state in experiment section. e.g., we set $p_\text{rl}$ as $0.5$ in all our experiments.}
%\UK{The detailed algorithm is presented in Appendix \ref{apx:network}.}

\subsection{Fitness-based Mean Update}
\label{sec:MU}
%\subsubsection{Rank-based Mean Update}
%\label{sec:RBMU}
%However,~\cite{ANES2013} modified rank-based update rule for the asynchronous scheme, by preserving the individuals of past $n$ updates. 
%, as in \eqnref{eq:asyncmu}. 
%The covariance matrix can be calculated with the archived individuals, previously described in \eqnref{eq:simplecovalcalc}. 
%\begin{equation}
%\mu_{t+1} = \mu_t + \frac{\eta}{n} \left(\sum_{i=1}^{K_e} \lambda_i z_i - \mu_t\right)
%\label{eq:asyncmu}
%\end{equation}
%We tested two variants in rank-based update methods.
%The first one is the same as the described method, which eliminates the \textit{oldest} one. 
%The second variant removes the \textit{worst} individual among the archived groups~\cite{selection1995}.
%We compare this method to our fitness-based update methods, which will be explained in the following, to show that fitness-based rule is more reliable in asynchronous algorithms.
%\subsubsection{Fitness-based Mean Update} 
%\label{sec:FBMU}
%To leverage the information from superior individuals maximally, we investigated a fitness-based method. 
%As introduced in \secref{sec:updates}, rank-based methods are more suitable for a synchronous update.
%Therefore, we design the smoothed update ratio $p$ between the two parameters, current $\mu$ and the newcomer.
%After the ratio is determined, the mean of the population is moved according to the equation \eqnref{eq:muupdate},
In rank-based methods, the valid weights of individuals for updating distribution are solely based on the rank, therefore the amount of ``how much better is the agent'' is ignored.
To address this problem, we consider fitness-based methods.
Fitness-based methods use the fitness value itself rather than rank, making it available to fully utilize the superior individual.
One of the early works, (1+1)-ES~\cite{Rechenberg1973}, compares the fitness values of a current and new individual. 
Although it is not considered as a fitness-based method, we can consider it as an extremely aggressive fitness-based method, since it moves the mean $\mu$ entirely to the new individual when the fitness value of the new one is better than the current one.
For smoother update, we design the mean update algorithm as~\eqnref{eq:muupdate},
\begin{equation}
\label{eq:muupdate}
    \mu_{t+1} = (1-p) \mu_t + p \cdot \mathbf{z}
\end{equation}
where $p$ is an update ratio between two parameters, and $\mathbf{z}$ is a parameter of newly evaluated individual.
Regarding~\eqnref{eq:muupdate}, $p$ in (1+1)-ES can be expressed as
\begin{equation}
    p = 
    \begin{cases} 
    1 & \text{if } f(\mu_t) < f(\mathbf{z}) \\
    0 & \text{if } f(\mu_t) \ge f(\mathbf{z})
    \end{cases}
\end{equation}
%When $p=0$, there occurs no update. 
%Similarly, the parameter is totally changed to new offspring when $p=1$.
%Otherwise, it updates softly and $p$ means the contribution of the parent and the offspring. 
%We discuss about the method for determining this update factor $p$ in the rest of this section. 
%A modified version of rank-based rule was introduced by~\cite{ANES2013}, which preserves the individuals of past $n$ updates.
%It removes the oldest individual when a new individual is evaluated, making it possible to apply the rank-based update rule in the same way to the asynchronous algorithm.
%However, learning rate $\eta$ should be $n$ times smaller since the update occurs $n$ times more frequently.
% \BU{clarify which p is novel and which is not}

In order to restrain the aggressiveness of (1+1)-ES, while maintaining the capability of obtaining much information from superior individuals, we propose two novel fitness-based update methods. 
% In order to restrain the aggressiveness of (1+1)-ES, while preserving the capability of high exploration in fitness-based scheme, we propose two novel fitness-based update methods.

\paragraph{Fixed Range}
First one is the fixed range method. 
This method updates the distribution relative to the current fitness value.
The update factor is determined by the fixed range around the fitness value of current mean. 
We evaluate two variants of the fixed range methods, it sets the update rate by clipped linear interpolation over a fixed range hyperparameter $r$. 
\begin{equation}
p = p_{\text{sg}} \cdot \text{clip}\left(\frac{f(\mathbf{z}) - f(\mu_t)}{r}, -1, 1 \right)
\end{equation}
$p_{\text{sg}}$ is determined as follow, 
\begin{equation}
    p_{\text{sg}} = 
    \begin{cases}
    p_{\text{positive}} & \text{if } f(\mu_t) < f(\mathbf{z}) \\
    p_{\text{negative}} & \text{if } f(\mu_t) \ge f(\mathbf{z})
    \end{cases}
\label{eq:psg}
\end{equation}
where $p_{\text{positive}}$ and $p_{\text{negative}}$ are hyperparameters.
%We can neglect the negative update by setting the $p_{\text{negative}}$ value, or different update factor can be applied between positive and negative udpates.
Note that it is possible to move $\mu$ only in positive direction by setting $p_{negative}$ as 0, or to let it move backward by setting $p_{negative}$ other than 0.

Additionally, we introduce Sigmoid function for more smooth update:
\begin{equation}
    p = p_{\text{sg}}\cdot \left[ \text{Sigmoid}\left( \frac{f(\mathbf{z}) - f(\mu_t)}{r}\right)\right]
\end{equation}
where $p_{\text{sg}}$ is the same as in~\eqnref{eq:psg}, and $\text{Sigmoid}(x) = 1/(1+e^{-x})$

\paragraph{Fitness Baseline} 
Secondly, we propose the baseline methods for more conservative updates compare to the fixed range methods. 
In the fixed range methods, the update ratio changes drastically with moving averages.
Therefore, we apply baseline techniques to make it less optimistic, as numerous RL methods~\cite{a3c16, td32018} advantage from.
There are two variants in this category: absolute baseline and relative baseline. 
The absolute baseline sets the baseline for the entire steps, while the relative method moves the baseline with the current mean value.
We also clip the update ratio from $-1$ to $1$. 
The former method follows the update rule in \eqnref{eq:absbase},
\begin{equation}
\label{eq:absbase}
    p = \text{clip}\left(\frac{f(\mathbf{z}) - f_{\text{b}}}{\left[(f(\mu) - f_{\text{b}}) + (f(\mathbf{z}) - f_{\text{b}})\right]}\right)
\end{equation}
where $f_{\text{b}}$ is a hyperparameter for the fitness baseline. 

The relative baseline is defined by $f_{\text{rb}} = f(\mu) - f_{\text{b}}$, which makes an update ratio $p$ and a modified version of $p_{\text{sg}}$ to be expressed as
\begin{equation}
    p = p_{\text{sg}} \cdot \text{clip} \left(\frac{f(\mathbf{z}) - f_{\text{rb}}}{f_b + (f(\mathbf{z}) - f_{\text{rb}})}, -1 ,1 \right), \qquad p_{\text{sg}} = 
    \begin{cases}
    p_{\text{positive}} & \text{if } f(\mathbf{z}) \ge f_{rb} \\
    p_{\text{negative}} & \text{if } f(\mathbf{z}) < f_{rb}
    \end{cases}
\end{equation}
We set $p=0$ for very low fitness value, $f(\mathbf{z}) < f_{rb} - f_b$
\subsection{Variance Matrix Update}
\label{sec:VMU}
We update the variance matrix $\mathbf{\sigma}I$ instead of the covariance matrix $\mathbf{\Sigma}$ to reduce time complexity, following the assumption used in CEM-RL~\cite{pourchot2018cemrl}.
As described in~\secref{sec:updates}, the variance of the synchronous method can be calculated directly from a set of individuals, but not in the asynchronous update method.
Therefore, different methods such as asynchronous ES~\cite{ANES2013} and Rechenberg's 1/5th success rule~\cite{Rechenberg1973} were introduced, where the first one utilizes pre-stored individuals, and the second one increases or decreases variance based on a success ratio $p_s$ and threshold ratio $p_{th}$.
However, the former simply adapted the method of rank-based synchronous, and the latter naively controlled the variance. 
In this section, we propose two asynchronous variance update methods that effectively encourages exploration in proportion to the update ratio of $p$. %Further, we introduce two online update algorithm inspired from the work~\cite{Welford62noteon}.

\paragraph{Online Update with Fixed Population}
In the asynchronous update scheme, the mean and variance of all individuals are unknown, and only the values of the current population distribution and a single individual are known. Welford's online update rule~\cite{Welford62noteon} is an update method that can be used in this situation, described as follows:

%\UK{In the asynchronous environment, the mean and variance of all individuals are unknown, and only the values of the current population distribution and a single individual are known.
%Welford's online update rule~\cite{Welford62noteon} is an update method that can be used in this situation, described as follows:
%}
%An online update rule by~\citet{Welford62noteon} is modified for variance updates. 
%The original online update rule is following:
\begin{equation}
    \mathbf{\sigma_t}^2 = \mathbf{\sigma_{t-1}}^2 + \frac{(\mathbf{z} - \mathbf{\mu_{t-1}})^\intercal(\mathbf{z} - \mathbf{\mu_t}) - \mathbf{\sigma_{t-1}}^2 }{n}
    \label{eq:welford}
\end{equation}
This equation was initially designed to save the memory space without storing the past data by calculating the new variance from the current variance. For this purpose, the original $n$ in \eqnref{eq:welford} is increased each time new data comes in, to represent the total number of data.
Also, past data and present data are equally valuable in the original equation.
However, since the past and present individuals should not be treated equally in the ES, we need to calculate the variance by giving more weight to the values of the recent individuals.
Thus, we modified $n$ in the equation to have constant value inspired by the concept of rank-based variance update method~\cite{ANES2013} which removes the oldest individual from the population.
The influence of old individuals would gradually disappear with the fixed $n$.
%Therefore, the original $n$ in \eqnref{eq:welford} indicates the total number of the data including a new one. 
%\UK{
%This equation was initially designed to save the memory space without storing the past data by calculating the new variance from the current %variance.For this purpose, the original $n$ in \eqnref{eq:welford} is added each time new data comes in and represents the total number of data.
%Also, past data and present data are equally valuable in the original equation.
%However, since the worth of past and present individuals are not equal in the ES and DRL fields, we need to modify them to calculate the variance by weighting the values of the current individual.
%Thus, we modified $n$ of the equation to have constant value inspired by the concept of rank-based variance update method~\cite{ANES2013} that remove the oldest individual from the population.
%We name this modified equation as an online update rule with a fixed population.
%}
%\UK{require expected benefit.}
%However, we modified this to gradually fade the influence of the old data.
%Two variants are tested to determine the $n$ value.
%The first is the fixed population method, simply set the $n$ to the constant value, similar to the concept that oldest individual is removed from the population. 

\paragraph{Online Update with Adaptive Population}
%In the adaptive population update rule, we utilize $p$ used in the mean calculation process of~\secref{sec:MU} to encourage exploration more effectively, in proportion to $p$.
%We utilize $p$ to estimate the importance of the new individual. 
%Therefore, the value of $n$ is estimated in reverse order from the update ratio $p$.
%The concept is inspired from the Rechenberg's method~\cite{Rechenberg1973}; its variance increases for better exploration when success rate is high, and decreases for more stability when success rate is low. 
%When $n$ is small, it has the effect that the previous variance is calculated from a few values.
%When $n$ is small, previous variance is calculated with only few values. 
%Therefore the influence of the new individual is larger.
%Similarly, when $n$ is large, we give more weights to the previous variance, thus the new individual affects less. 
%For example, let the $p$ value is $0.25$. 
%To make the $p$ value from a new individual, $3$ old individuals should be laid on $0$.
%Then, we clipped the minimum of $n$ to $1$.
%Resulting equation is following:
In the fixed population update rule, the number of the past individuals $n$ is fixed during the update.
It means that every new individual added to the calculation will have the same influence in every update.
To make this process more adaptive to each individual, we consider Rechenberg's method~\cite{Rechenberg1973}.
It makes its variance increase for better exploration when success rate is high, and decrease for more stability when success rate is low.

We design $n$ by utilizing the update ratio $p$ from the mean calculation in~\secref{sec:MU}, to replace `success rate' in~\cite{Rechenberg1973} to be more suitable for our algorithm, as well as to smooth and stabilize the update process.
$p$ implies how `important' the new individual is.

To elaborate, assume $p$ of the newly evaluated individual is high.
This means that the individual is relatively `important' than the other individuals.
We want to increase the influence of this individual in the variance update, by reducing the value of $n$, which reduces the influence of the previous variance $\sigma_{t-1}$ in~\eqnref{eq:welford}.
If the new individual is relatively not `important', i.e. it has small $p$, then $n$ is set to a larger value. When $p=0$, then the new individual has no importance, the variance remains unchanged. 
With a clipping operation added, we can express the update process of $n$ as
\begin{equation}
    n = \max\left(\frac{1-p}{p}, 1\right)
    \label{eq:AdaptivePopulation}
\end{equation}
% \vspace{-0.1in}

%\UK{In the adaptive population update rule, we utilize the $p$ used in the mean calculation process of~\secref{sec:MU} to more effectively encourage exploration in proportion to $p$.} 
%The second is the adaptive population method. 
%\UK{The size of the population $n$ can be estimated inversely from the update ratio $p$.}
%The size of the population is estimated in reverse order from the update ratio $p$.
%That is, $n$ here means that "how many individuals were $0$" if the new mean is moved to $p$ when a new individual of value $1$ arrived.
%For example, let the $p$ value is $0.25$. To make the $p$ value from a new individual that has the value $1$, $3$ individuals should be laid on $0$.
%\UK{where the large update ratio $p$ means that the fitness or rank of the corresponding individual is high, so the variances are updated a lot.
%Conversely, when the update ratio $p$ is small, the variances are updated less.
%}
%We suggested various update rules for the asynchronous method along with the previously proposed algorithms. 
% We evaluated all combination in \secref{sec:compareproposed}
%We evaluated all of the combinations, the results are presented in Appendix \ref{apx:result}. 
% The detailed derivation of~\eqnref{eq:welford} and~\eqnref{eq:AdaptivePopulation} are presented in Appendix \ref{apx:DAP}. 
Note that the mean and the variance update rules are independent, so any combination of the rules can be used. 

\section{Experiment}
\label{sec:experiment}
In this section, we compare the results of provided methods with the synchronous algorithm in the perspective of time-efficiency and performance. 
We evaluate the algorithms in several simulated environments which are commonly used as benchmarks in policy search: HalfCheetah-v2, Hopper-v2, Walker2d-v2, Swimmer-v2, Ant-v2, and Humanoid-v2~\cite{mujoco2018}.
The presented statistics were calculated and averaged over 10 runs with the same configuration. 
As our aim is to show the efficiency of the asynchronous methods, we tried to use architecture and hyperparameters as similar as possible of those in CEM-RL~\cite{pourchot2018cemrl}.
We used TD3 \cite{td32018} in RL part. 
Detailed architecture and hyperparameters for all methods are shown in Appendix \ref{apx:network} and \ref{apx:hyper}, respectively. \footnote{The source code of our implementation is available at \url{https://github.com/KyunghyunLee/aes-rl}}

\subsection{Time Efficiency Analysis}
We compare the time efficiency of three architectures, serial-synchronous, parallel-synchronous, and parallel-asynchronous. 
CEM-RL from the original author\footnote{\url{https://github.com/apourchot/CEM-RL}} is based on a serial-synchronous method. We implement parallel-synchronous version of CEM-RL, P-CEM-RL. 
Finally, AES-RL is our proposed parallel-asynchronous method. 
%All three algorithms are evaluated in HalfCHeetah-v2, which has the fixed episode steps and Hopper-v2, which has a varying episode steps, with the same hardware configuration that AMD Ryzen 3900x CPU, 2 $\times$ NVidia 2080Ti GPUs, and five workers.
All three algorithms are evaluated in HalfCHeetah-v2, which has the fixed episode steps, and Walker2d-v2 and Hopper-v2, which has varying episode steps, in the same hardware configuration, two Ethernet-connected machines of Intel i7-6800k and three NVidia GeForce 1080Ti; a total of 24 CPU cores and 6 GPUs.
% 2 worker situation can show the boosting result. 

The results are shown in \tabref{tbl:timeresult}.
It shows that parallelization brings significant time reduction compared to the serial methods.
%Simply switching from serial to parallel, time efficiency is greatly improved, 63\% of total training time is reduced compared to the synchronous serial method in HalfCheetah-v2.
%Moreover, additional efficiency is achieved from asynchronism. Consequently, 66\% is reduced.
By simply switching from serial to parallel, time efficiency is greatly improved, reducing around 60\% of total training time compared to the synchronous serial method. Additional efficiency of around 10-20\% is achieved from asynchronism.
We also scale workers up to nine, then 80-90\% of the training time is reduced compared to the serial version.
%In Hopper-v2, the total training time is reduced 48\% with parallel-synchronous method, further 61\% is reduced with the asynchronous method.
The result shows that the inefficiency of the synchronous methods is significant with varying episode length, as described in \figref{fig:overview}.
% The improvement is more significant in Hopper-v2 because the episode length differs.

% \BU{Due to the limited hardware resource, we could not scale up to dozens of workers, we can expect the efficiency difference will be increased \cite{Martin2015AsynchronousPE}. }

% \BU{little more explanation}

\begin{table}[t]
%  \caption{{\bf Time efficiency comparison of three algorithms evaluated in HalfCheetah-v2 and Hopper-v2.} In HalfCheetah-v2, the episode length is fixed to 1000 steps, evaluation time for all agents are theoretically identical. In Hopper-v2, the episode length varies according to the policy. }
  \caption{{\bf Time efficiency comparison results in HalfCheetah-v2, Walker2d-v2 and Hopper-v2.} 
  In HalfCheetah-v2, the episode length is fixed to 1000 steps, and evaluation time for all agents are theoretically identical. 
  In Walker2d-v2 and Hopper-v2, the episode length varies. 
  The percentage indicates the amount of \textit{reduced} time compared to CEM-RL. 
  More detailed results about the number of workers are provided in \adxref{apx:numberofworkers}. }
  \label{tbl:timeresult}
  \vspace{-0.05in}
  \centering
  \begin{tabular}{r|c|c|c|c}
    \toprule
     & CEM-RL & P-CEM-RL & \multicolumn{2}{c}{AES-RL} \\
    \hline
     Sequence & Serial & Parallel & \multicolumn{2}{c}{Parallel} \\
     Update & Sync. & Sync. & \multicolumn{2}{c}{Async.} \\
     \hline
     Workers & 1 & 5 & 5 & 9 \\ 
    \midrule
    \midrule
    HalfCheetah-v2 (min) & 467 & 187 (59.9\%$\downarrow$) & 83 (82.1\%$\downarrow$) & 54 (88.3\%$\downarrow$)\\
    Walker2d-v2 (min) & 487 & 205 (57.9\%$\downarrow$) & 105 (78.4\%$\downarrow$) & 77 (84.1\%$\downarrow$)\\
    Hopper-v2 (min) & 505 & 188 (62.6\%$\downarrow$) & 133 (73.6\%$\downarrow$) & 91 (82.0\%$\downarrow$) \\
    \bottomrule
  \end{tabular}
  \vspace{-0.05in}
\end{table}

\begin{table}[t]
  \caption{{\bf Performance analysis for various combinations of our proposed mean-variance update methods.} 
  Results are measured with average scores of ten test runs within a total of 1M step from the summation of all worker steps, averaged with ten random seeds.
  Full results for various environments are presented in \adxref{apx:result}. 
   }
%  \caption{{\bf Partial comparison results of Mean-Variance update methods in HalfCheetah-v2 and Walker-v2.} Full results are presented in Appendix \ref{apx:result}}
%  \caption{{\bf Partial results of previous and proposed algorithms tested in HalfCheetah-v2 and Walker-v2.} Full results are presented in Appendix \ref{apx:result}}
  \vspace{-0.05in}
  \label{tbl:algorithmresult}
  \begin{adjustbox}{max width=\textwidth}
  \centering
  \begin{tabular}{l|c||c|c||c|c|c|c|c|c|c|c}
    \toprule
    \multirow{4}{*}{} & \multirow{2}{*}{Category} & \multicolumn{2}{c||}{Previous algorithms} & \multicolumn{8}{c}{Proposed algorithms} \\
    \cline{3-12}
       &  & (1+1)-ES & Rank-Based  &
    \multicolumn{2}{c|}{Fixed Range} & \multicolumn{2}{c|}{Fitness Baseline} & \multicolumn{2}{c|}{Fixed Range} & \multicolumn{2}{c}{Fitness Baseline} \\
    \cline{2-12}
    & $\mu$ & Full Move & \multirow{2}{*}{Oldest} & Linear & Sigmoid & Absolute & Relative & Linear & Sigmoid & Absolute & Relative \\
    \cline{2-3} \cline{5-12}
    & $\mathbf{\Sigma}$ & Success Rule &    & \multicolumn{4}{c|}{Online Update - Adaptive}& \multicolumn{4}{c}{Online Update - Fixed}  \\
    % \hline
    % \hline
    \midrule
    \multirow{2}{*}{HalfCheetah-v2} & Mean & 11882 & 10010 & 10279 & 12053 & 12224 & \textbf{12550} & 10870 & 12031 & 11767 & 12128  \\
                                    & Std. & 385   & 746   & 1044  & 398   & 422   & \textbf{187}   & 409   & 604   & 458   & 821   \\ 
    \hline
    \multirow{2}{*}{Walker2D-v2}  & Mean & 2347 & 4230 & 5020 & 5360 & 5137 & \textbf{5474} & 3419 & 5121 & 5039 & 5070 \\
                                  & Std. & 320  & 254  & 799  & 683  & 223  & \textbf{223}  & 1676 & 965    & 471    & 557    \\
    \bottomrule
  \end{tabular}
  \end{adjustbox}
  \vspace{-0.15in}
\end{table}

%\subsection{Performance Compare}
% In this section, we analyze the performance of our algorithms from two perspectives. 
% We evaluate all combination of our mean and variance update methods, in addition to the result of previous asynchronous algorithms~\cite{selection1995, ANES2013}.
%Our experiment consist of two parts. 
%Firstly, we evaluate various combination of algorithms proposed in \secref{sec:methods} and analyze it. 
%Secondly, we compare our method to the previous methods. 
\subsection{Performance Analysis of Mean-Variance Update Rules}
\label{sec:meanvarupdate}
%\subsection{Results of the Proposed Update Algorithms} 
We propose the asynchronous update methods including 4 mean and 2 variance update rules, as described in \secref{sec:methods}.
Therefore, 8 combinations are available, and we thoroughly analyze their properties.
%We examine the asynchronous update methods previously described in \secref{sec:methods}. 4 Methods for mean updates and 2 methods for variance updates are proposed.Therefore, 8 combinations are possible. 
\tabref{tbl:algorithmresult} shows the comparison results, along with the previously proposed algorithms, (1+1)-ES~\cite{Rechenberg1973}, and asynchronous version of rank-based update~\cite{ANES2013}. 
% All algorithms are evaluated in HalfCheetah-v2, relatively easy, and Walker-v2, a harder environment.
All algorithms are evaluated in HalfCheetah-v2 and Walker-v2 environment, where former is easier and latter is harder, relatively.

As we expected in \secref{sec:methods}, (1+1)-ES shows the instability because of its extreme update rule.
It successfully solved the HalfCheetah-v2, but failed in Walker-v2.
In contrast to the (1+1)-ES, Rank-based method shows lower performance in both environment. 
They saturated at the lower score and was not able to explore further. 

%Compared to the baseline algorithms, fixed range algorithms show lower performance, yet better than the rank-based method.
The fixed range algorithms show lower performance compared to the baseline algorithms, but are better than the rank-based method.
Sigmoid is better than linear for all configuration, since it leverages more advantages from superior individuals.
%Baseline methods show higher performance and stability.
%They are consistently better than other methods and shows lower variance in performance.
The baseline methods show higher performance and stability, showing higher mean and lower variance, consistently.
Here, stability is defined as how consistent the results are for various random seeds with the std value per mean $\sigma/\mu$. 
In the absolute baseline method, it updates conservatively at the latter part of training because of the scaling effect mentioned in \secref{sec:updates}.
Whereas, the relative baseline method effectively utilizes the information from superior individuals, thus showing the best results.
In terms of the variance update, the adaptive method shows better performance, compare to the fixed population method.
The fixed population method shows higher variance and lower mean, because it sometimes fails to explore enough compared to the adaptive method.
As a result, the relative baseline with adaptive variance method shows the best results as expected.
Detailed result for combinations are presented in Appendix~\ref{apx:result}.

% \vspace{-0.2in}
\begin{table}[b!]
  \caption{{\bf Performance analysis of TD3, SAC, CEM, ERL, CEM-RL, and AES-RL in six MuJoCo benchmarks.} Our algorithm outperforms the other methods in most of environments except for Ant-v2 and Swimmer-v2. Results of other algorithms are from their original report, except for Humanoid-v2. Improvements are compared to the baseline CEM-RL. }
%  \caption{Performance of TD3, CEM, ERL, CEM-RL, and AES-RL in five MuJoCo benchmarks. Our algorithm outperforms every other methods in all environments except for Swimmer-v2. It starts to solve Swimmer, but not always. }
  \vspace{-0.05in}
  \label{tbl:result}
  \centering
  \begin{adjustbox}{max width=\textwidth}
  \begin{tabular}{ll|cc|c|cc|c|c}
    \toprule
    Environment     & Statistics     & TD3 & SAC & CEM &  ERL & CEM-RL & AES-RL & Improvement\\
    \midrule
    \multicolumn{2}{l|}{Algorithm}&   \multicolumn{2}{c|}{RL} & ES      &   \multicolumn{2}{c|}{RL+ES}   &   RL+ES     &  \\
    \cline{3-8}
    \multicolumn{2}{l|}{Update Method} & \multicolumn{2}{c|}{-}  &  -      &   \multicolumn{2}{c|}{Sync.}   &   Async. &   \\
    \midrule
    \midrule
    \multirow{3}{*}{HalfCheetah-v2} &   Mean    & 9630 & 11504 &  2940    &   8684    &   10725   &   \textbf{12550}  & 
    \multirow{3}{*}{\textbf{17.02 \%}}\\
                                    &   Std.    & 202  & 183 &  353     &   130     &   397     &   \textbf{187}    \\
                                    &   Median  & 9606 & 11418 &  3045    &   8675    &   11539   &   \textbf{12571}  \\
    \hline
    \multirow{3}{*}{Hopper-v2}      &   Mean    & 3355 & 3239 &  1055    &   2288    &   3613    &   \textbf{3751}   & 
    \multirow{3}{*}{\textbf{3.83 \%}}\\
                                    &   Std.    & 171  & 18 &  14      &   240     &   105     &   \textbf{58}     \\
                                    &   Median  & 3626 & 3230 &  1040    &   2267    &   3722    &   \textbf{3746}   \\
    \hline
    \multirow{3}{*}{Walker2D-v2}    &   Mean    & 3808 & 4268 &  928     &   2188    &   4711    &   \textbf{5474}   & 
    \multirow{3}{*}{\textbf{16.19 \%}}\\
                                    &   Std.    & 339  & 435 &  50      &   328     &   155     &   \textbf{223}    \\
                                    &   Median  & 3882 & 4354 &  934     &   2338    &   4637    &   \textbf{5393}   \\
    \hline
    \multirow{3}{*}{Ant-v2}         &   Mean    & 4027 & \textbf{5985} &  487     &   3716    &   4251    &   5120   &
    \multirow{3}{*}{\textbf{20.44 \%}}\\
                                    &   Std.    & 403  & \textbf{114} &  33      &   673     &   251     &   170    \\
                                    &   Median  & 4587 & \textbf{6032} &  506     &4240    &   4310    &   5071   \\
    \hline
    \multirow{3}{*}{Swimmer-v2}     &   Mean    & 63   & 46 &  \textbf{351} &   350    &   75      &   161    & \multirow{3}{*}{-}\\
                                    &   Std.    & 9    & 2 &   \textbf{9}   &   8      &   11      &   100    \\
                                    &   Median  & 47   & 45 &  \textbf{361} &   360    &   62      &   128    \\
    \hline
    \multirow{3}{*}{Humanoid-v2}    &   Mean    & 5496 & 5505 &  -   &   5170   &   5579    &   \textbf{6136}   & \multirow{3}{*}{\textbf{9.98 \%}}\\
                                    &   Std.    & 187  & 108 &  -   &   130    &   228     &   \textbf{444}      \\
                                    &   Median  & 5465 & 5539 &  -   &   5120   &   5673    &   \textbf{6133}      \\
    % \hline
    % \multirow{3}{*}{Reacher-v2}     &   Mean    &   -       &   \textbf{-}   &   -      &   -       &   -      & \multirow{3}{*}{-}\\
    %                                 &   Std.    &   -       &   \textbf{-}   &   -      &   -       &   -      \\
    %                                 &   Median  &   -       &   \textbf{-}   &   -      &   -       &   -      \\

    \bottomrule
  \end{tabular}
  \end{adjustbox}
  \vspace{-0.1in}
\end{table}
\subsection{Performance Analysis of Other Algorithms}
%\subsection{Compare with Previous Algorithms}
We compare our proposed algorithm to the previous RL and ERL algorithms: TD3, SAC~\cite{haarnoja2018soft}, CEM, ERL and CEM-RL.
%\BU{brief info of each algorithms? in background section 2.4}
The results are summarized in \tabref{tbl:result}.
Except for our results, we cite the results of the RL and ERL algorithms reported in~\cite{pourchot2018cemrl}.
Excluding the ERL, the other algorithms are trained for 1M steps.
Training steps of the ERL is the same as what were given in the original paper~\cite{erl2018}: 2M on HalfCheetah-v2, Swimmer-v2 and Humanoid-v2, 4M on Hopper-v2, 6M on Ant-v2 and 10M on Walker2d-v2. 
%The values of algorithms other than ours are as reported in~\cite{pourchot2018cemrl}.
%Excluding the ERL, the result values of the other algorithms are trained for 1M steps. 

Our algorithm outperforms most of the other methods consistently, except for Ant-v2 and Swimmer-v2.
As reported in CEM-RL, Swimmer-v2 is a challenging environment for all RL algorithms. 
However, our methods show better performance than TD3, CEM-RL in Swimmer-v2, because of better exploration. 
%In detail, our algorithm gets high-score 2 out of 10 trials, and low-score 8 out of 10 trials, in contrast to CEM-RL and TD3 that get low-score all trials.
In detail, our algorithm gets two high scores and eight low scores out of 10 trials, in contrast to CEM-RL and TD3 that got low scores for all trials.
The results for failed trials are \textbf{111.97 $\pm$ 24.10}, and \textbf{355.95 $\pm$ 1.95} for the succeeded trials. 
%With this perspective, our method improved the CEM-RL in failed attempts, and it is noticeable that the algorithm starts to solve the environment. 
%This can be evidence of the exploration boosting of our methods, on the border of success and failure.
This shows that even in our 8 failed attempts, we outscore CEM-RL, while showing the comparable results in 2 succeeded trials.

\begin{figure}[h!]
\caption{\textbf{Wallclock-wise Learning Curve} Curves are averaged with three random seed}
\label{fig:wallclocklearning}
\centering
\begin{adjustbox}{max width=\textwidth}
\begin{tabular}{cc}
    % \vspace{-0.2in}
    \includegraphics{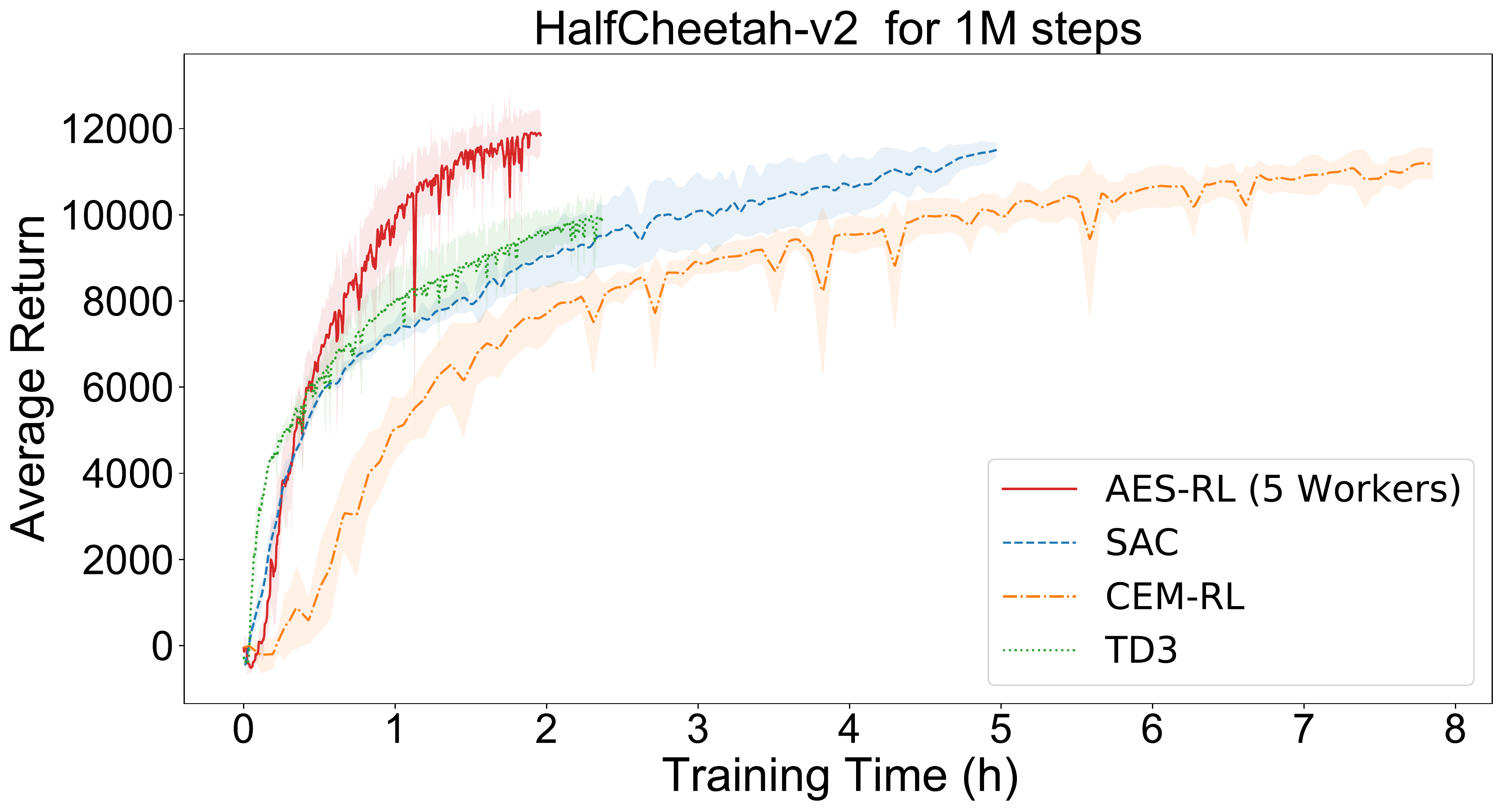} &
    \includegraphics{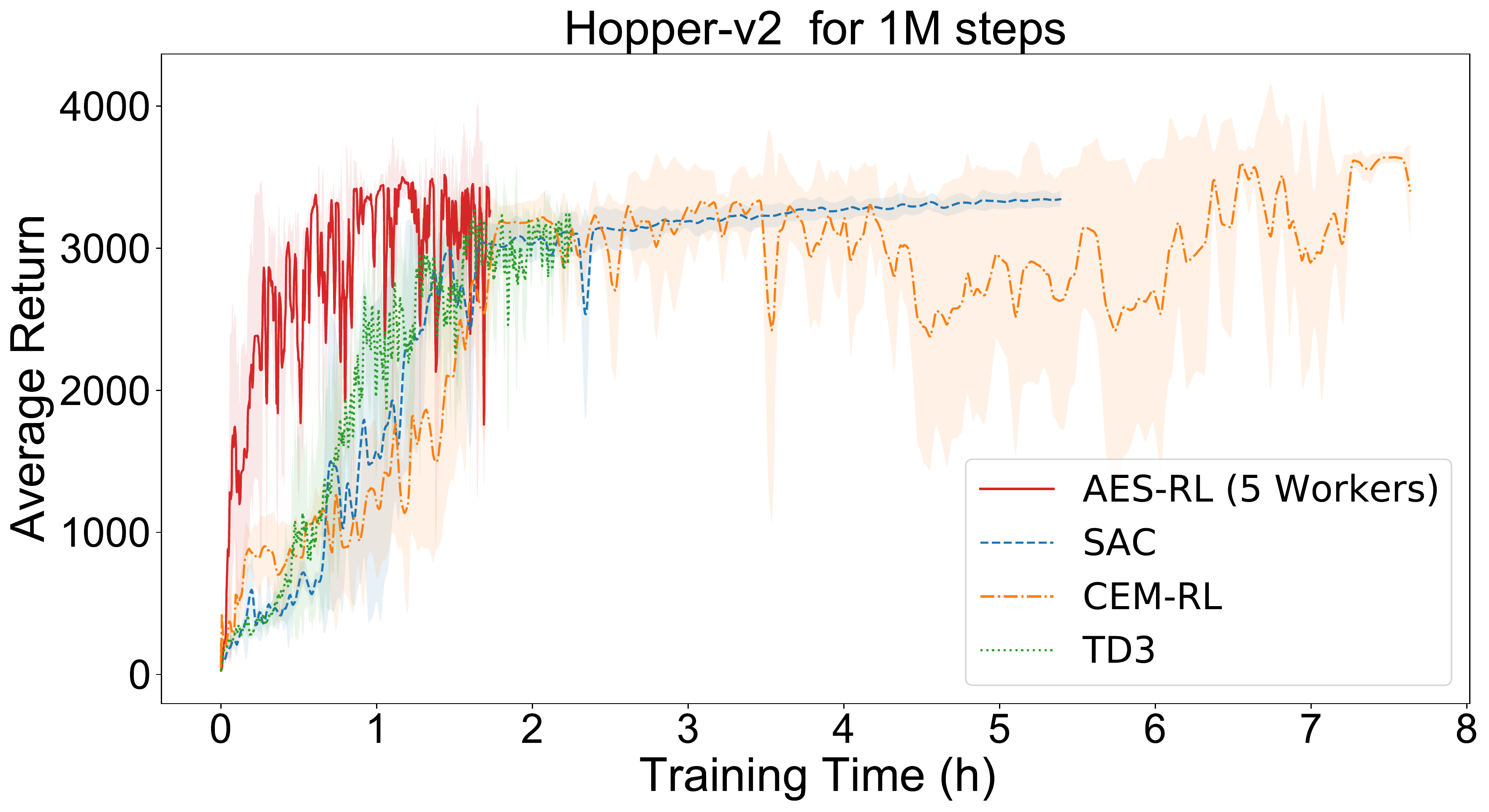} \\
    \includegraphics{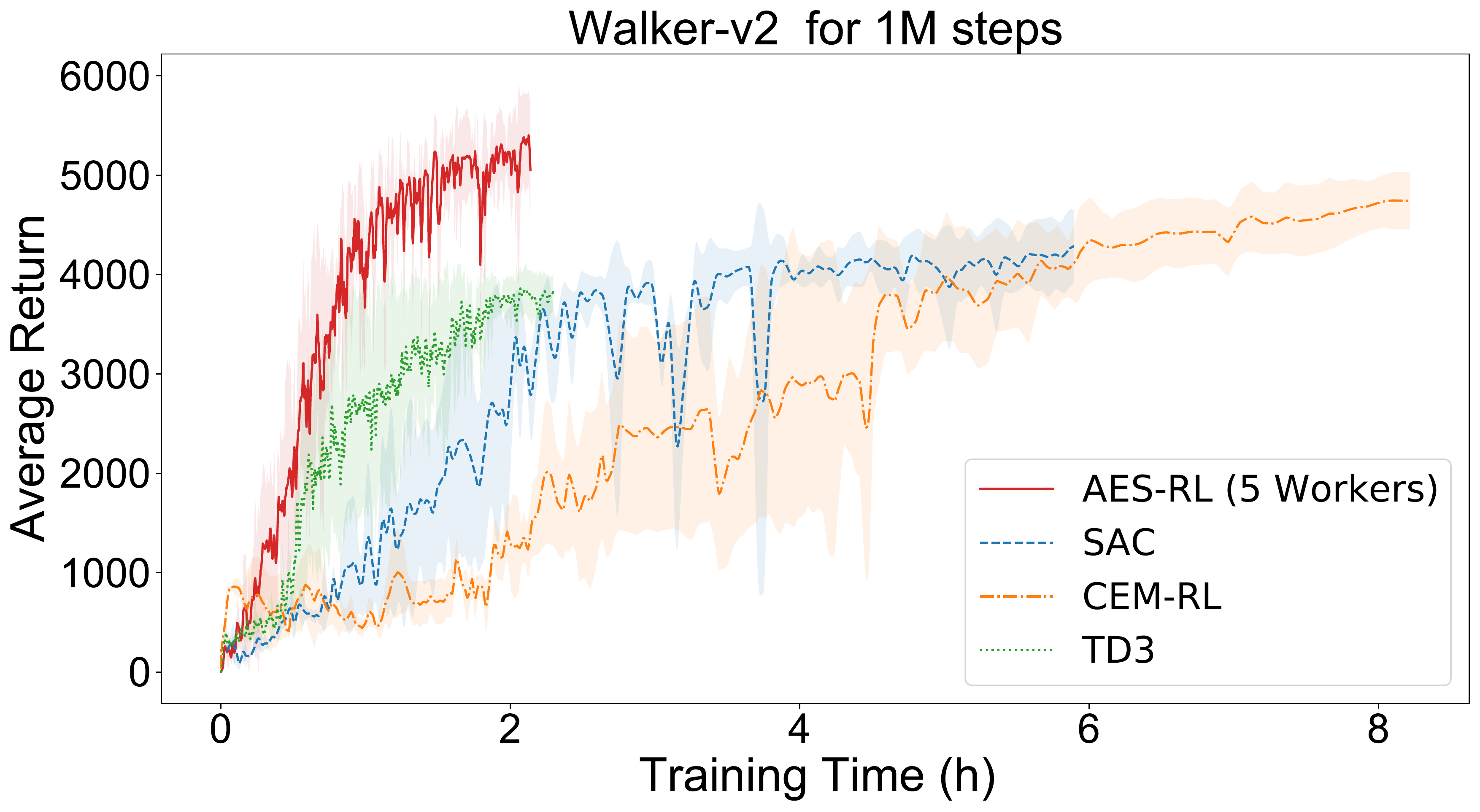} &
    \includegraphics{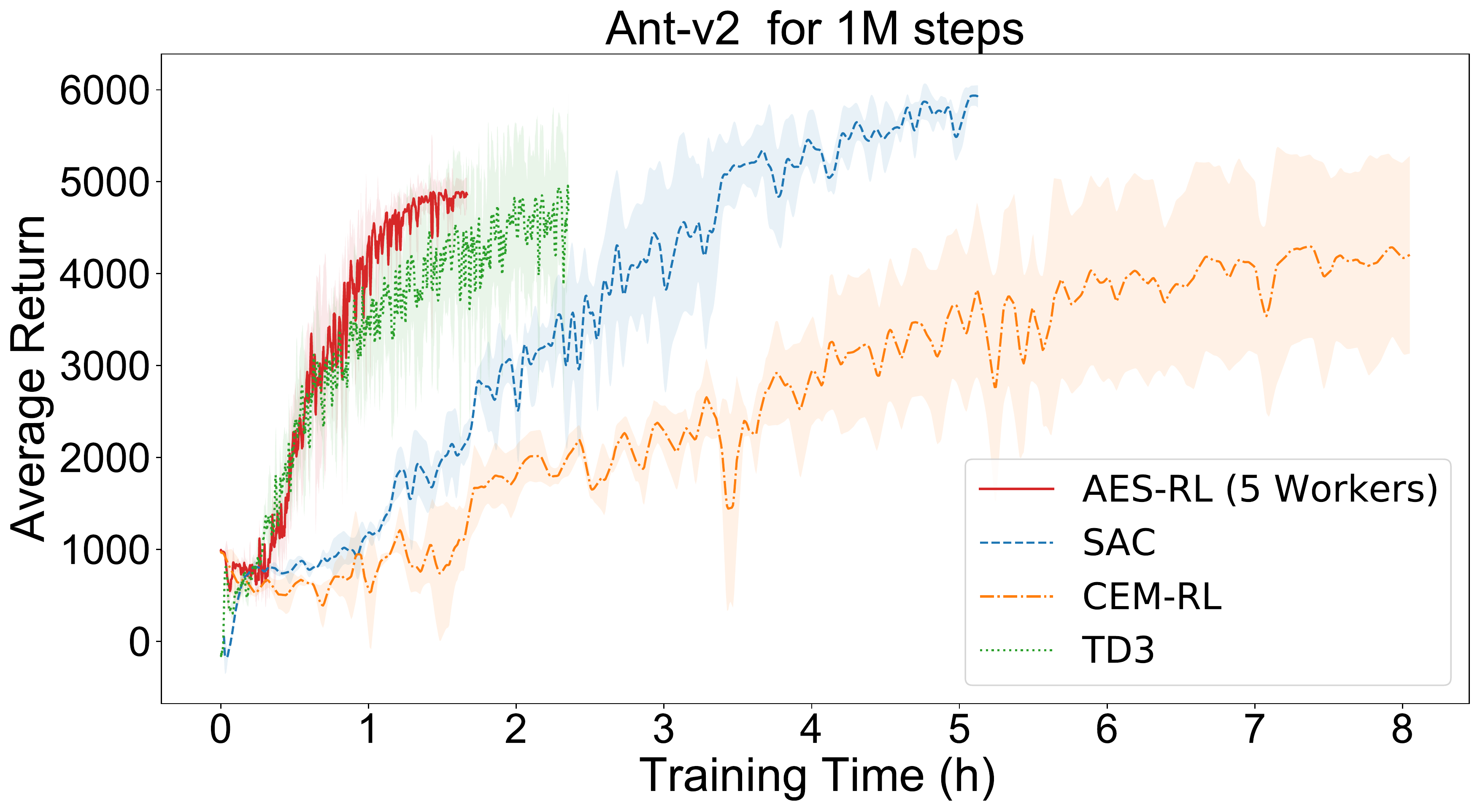} \\
    \includegraphics{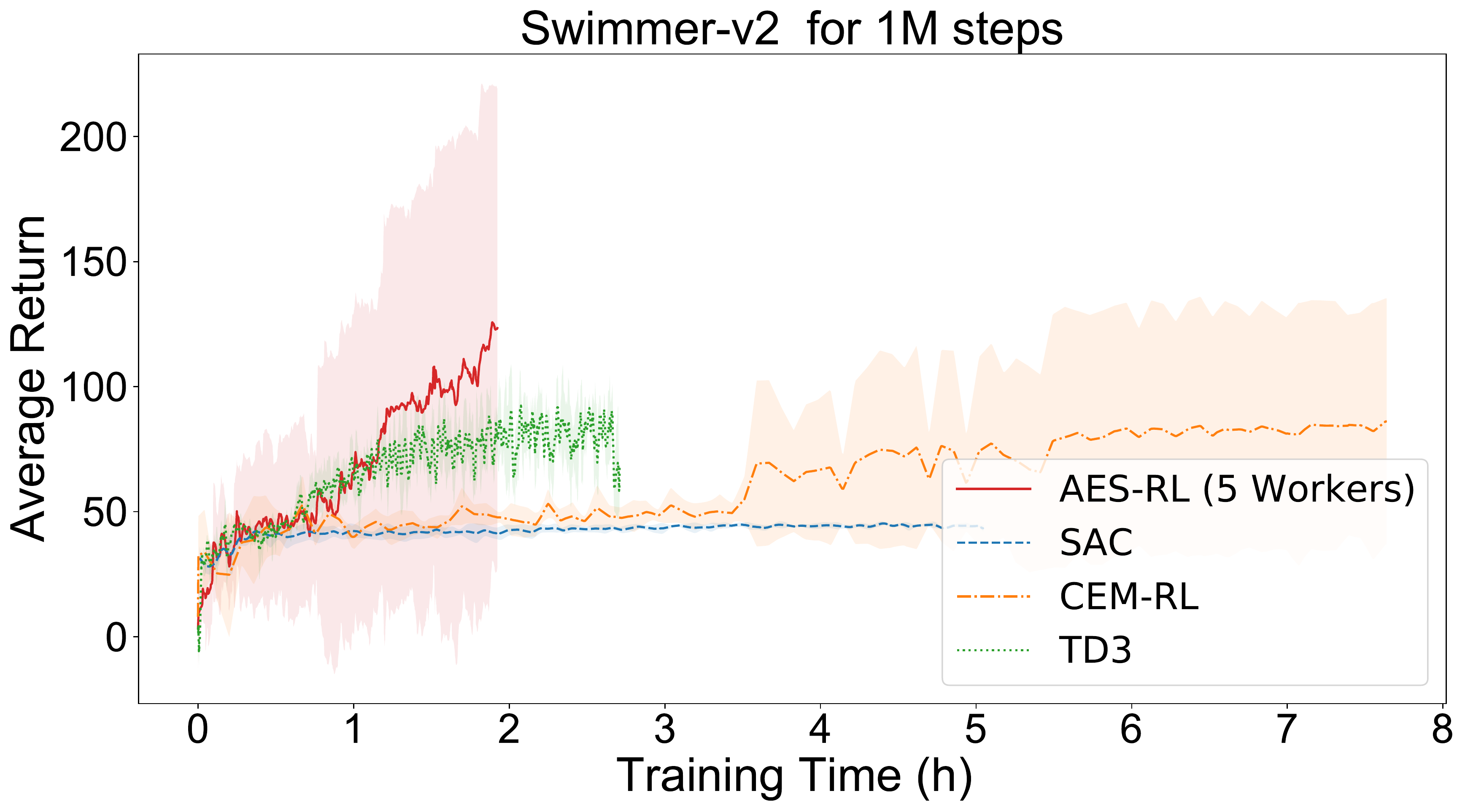} &
    \includegraphics{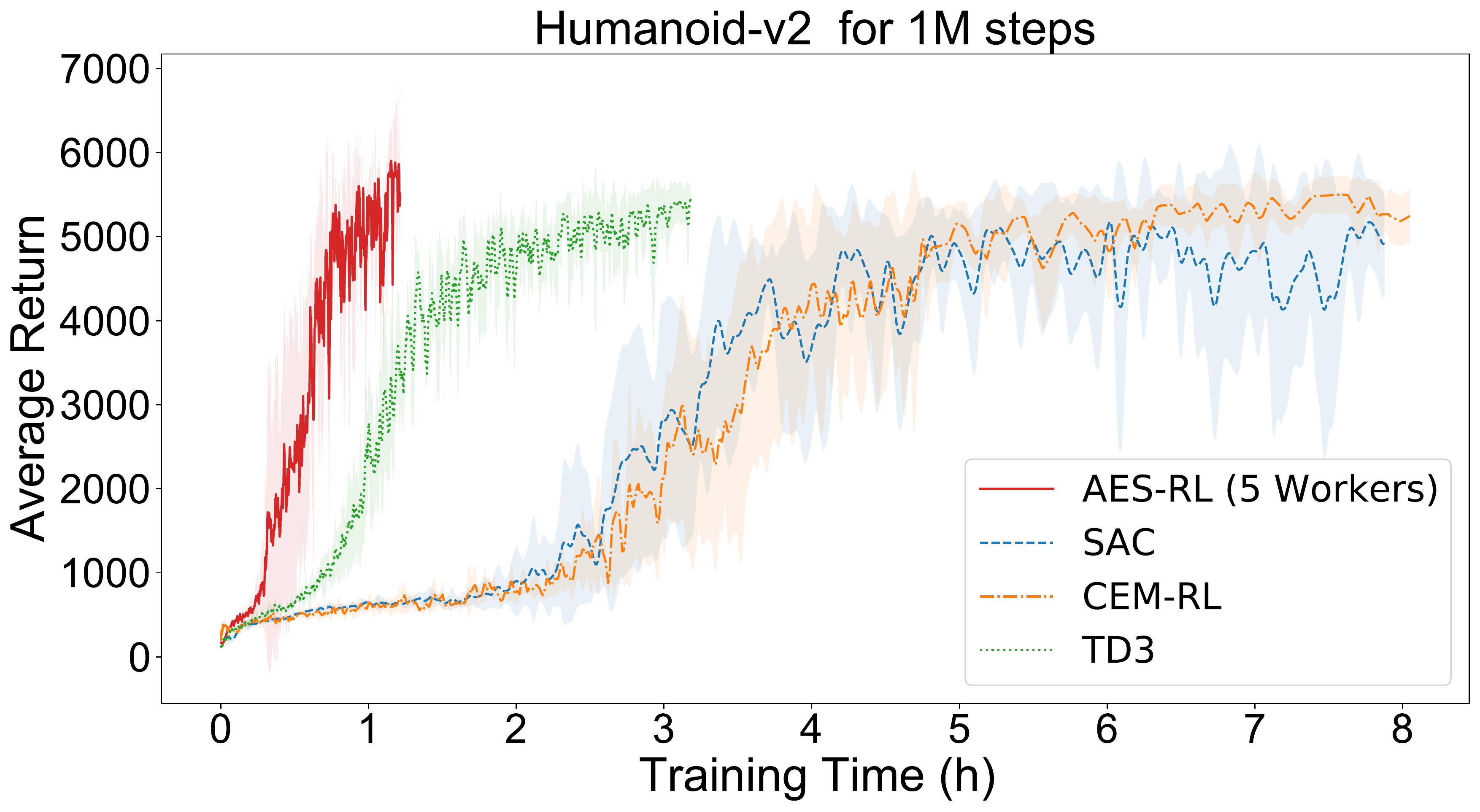} \\
    \vspace{-0.2in}
\end{tabular} 
\end{adjustbox}
\end{figure}

The learning curves for each environment are presented in~\figref{fig:wallclocklearning}. 
It shows that the AES-RL can achieve better performance with less time and also within a fixed amount of steps.

%As in CEM-RL, our methods are not stable for Swimmer-v2. 
%However, the algorithm solves the environment twice out of ten times, and fails eight out of ten times, in contrast to CEM-RL and TD3 that failed all trials. 
% For Swimmer-v2, conventional RL algorithms stuck to the performance of $\sim 130$ \cite{poplin2019}. 
% Our method fails to solve the Swimmer-v2 stably. 
% However, it sometimes overcomes the local minima, unlike the CEM-RL. 

% \begin{figure}[t]
%   \centering
%   \fbox{\rule[-.5cm]{0cm}{4cm} \rule[-.5cm]{12cm}{0cm}}
%   \caption{\todo{Learning Cruve Graph}}
% \end{figure}

\section{Discussion}
In this paper, we proposed the asynchronous parallel ERL framework to address the problems of previous synchronous ERL methods. 
We also provided several novel mean-variance update rules for updating the distribution of the population and analyzed the influence of each method. 
Our proposed framework and update rules were evaluated in six MuJoCo benchmarks, and showed outstanding results with 80\% of time reduction with five workers and 20\% of performance improvement at its best.
%Our proposed framework and update rules are evaluated in five MuJoCo benchmarks, and shows its time efficiency and performance improvements, almost 60\% time reduction and 20\% improvement except for Swimmer-v2, respectively. 
%For Swimmer-v2, although there was no performance improvement compared to CEM and ERL, it showed a significant performance improvement over CEM-RL, our baseline, and sometimes it was comparable to CEM and ERL.
% For Swimmer-v2, although the averaged statistics indicated no improvement, we were able to prove that our algorithm has the capability of high exploration that is far beyond CEM-RL.

With our framework, it is possible to use any off-policy RL algorithms and many ask-and-tell ES algorithms, which can be used with up to 200k parameters.
We remain this for the future work.  

\section*{Broader Impact}
To apply RL in the real world problem, efficiency and stability are essential, because the cost of interacting with the environment is much higher than the simulation. 
Evolutionary Reinforcement Learning is an attempt to combine ES and RL for efficiency and stability.
However, the previous works used synchronous evolutionary algorithms that cannot be scaled up.

This paper focuses on the asynchronous combination method of RL and ES. 
We introduce several efficient update rules for applying asynchronism and analyze the performance.
We confirm that the asynchronous methods are much more time efficient, and it also have more exploration property for searching effective policy. 
With the proposed algorithm, AES-RL, branch of stable and efficient policy search algorithms can be extended to numerous workers.
We expect the policy search algorithms to be actively applied to real world problems.

\begin{ack}
This work was in partial supported by the Industrial Strategic Technology Development Program(Development of core technology for advanced locomotion/manipulation based on high-speed/power robot platform and robot intelligence, 1 0070171) funded By the Ministry of Trade, industry \& Energy(MI, Korea)
This work was also in partial supported by the Technology Innovation Program funded by the Ministry of Trade, Industry and Energy, South Korea, under Grant 2017-10069072.
\end{ack}

\bibliographystyle{unsrtnat}
\bibliography{reference}
\medskip

\small

\newpage
\appendix
% \LARGE{\textbf{Appendix}}
\normalsize
\section{Network Architecture}
\label{apx:network}
For a fair comparison, our network follows the same structure as CEM-RL~\cite{pourchot2018cemrl}. The architecture is originally from \citet{td32018}, the only difference is using $\text{tanh}$ instead of $\text{RELU}$. \citet{pourchot2018cemrl} reported the difference between the $\text{RELU}$ and $\text{tanh}$. We use $(400, 300)$ hidden layer for all environment except Humanoid-v2. For Humanoid-v2, we used $(256, 256)$ as in TD3~\cite{td32018}.

\begin{table}[h!]
%  \caption{{\bf Time efficiency comparison of three algorithms evaluated in HalfCheetah-v2 and Hopper-v2.} In HalfCheetah-v2, the episode length is fixed to 1000 steps, evaluation time for all agents are theoretically identical. In Hopper-v2, the episode length varies according to the policy. }
  \caption{{\bf Network architectures} The architecture from the input layer to the output layer}
  \label{tbl:architecture}
  \vspace{-0.1in}
  \centering
  \begin{tabular}{l|cc}
    \toprule
    Layer Type& Actor &  Critic \\
    \hline
    Linear      & (state\_dim, 400)     &  (state\_dim + action\_dim , 400) \\
    Activation  & tanh                  &   leaky RELU \\
    Linear      & (400, 300)            &   (400, 300) \\
    Activation  & tanh                  &   leaky RELU \\
    Linear      & (300, action\_dim)    &   (300, 1) \\
    Activation  & tanh                  &           \\
    \bottomrule
  \end{tabular}
  \vspace{-0.2in}
\end{table}

\section{AES-RL pseudo-code}
\label{apx:algorithm}
% Our original source code will be released through github as soon as possible. 

\begin{algorithm}[h!]
\DontPrintSemicolon
\caption{AES-RL}
\textbf{Initialize: } the mean of the population $\pi_\mu$, shared critic $Q^\pi$ and target critic ${Q^\pi}'$ \\
\textbf{Initialize: } the covariance matrix $\mathbf{\Sigma} = \sigma_{\text{init}} \mathcal{I}$, the emtpy replay buffer $\mathcal{R}$, $total\_steps = 0$ \\
\tcc{Start training of the shared critic}
critic\_worker.start\_critic\_training() 

\While{$total\_steps < max\_steps$}{
    $N_{idle\_worker} = $ num\_idle\_worker() \\

    \While {$N_{idle\_worker}$ > 0} {
        \tcc{Create new individual} 
        new\_individual = population.sample($\pi_\mu$, $\mathbf{\Sigma}$) \\
        Create new individual by sampling from $\mathcal{N}(\pi_\mu, \mathbf{\Sigma})$ \\
        actor\_worker = get\_idle\_worker() \\
        actor\_worker.set\_actor\_network(new\_individual) \\
        Calculate $p_{\text{rl}}$ according to \eqnref{eq:prl} \\
        \uIf{$\text{rand}() < p_{\text{rl}}$ {\bf and} total\_step >= rl\_start\_step}{
            \tcc{Train the actor with shared critic}
            critic\_weight = critic\_worker.get\_critic\_weight() \\
            actor\_worker.set\_critic\_network(critic\_weight) \\
            actor\_worker.train\_actor\_network() \\
            $n_{\text{rl}} \leftarrow n_{\text{rl}} + 1$ \\
        }
        \Else{
            \tcc{Evaluate immediately}
            $n_{\text{es}} \leftarrow n_{\text{es}} + 1$ \\
        }
        \tcc{Evaluate the individual, with filling the replay buffer}
        actor\_worker.evaluate($\mathcal{R}$)\\
        $N_{idle\_worker} \leftarrow N_{idle\_worker} - 1$
    }
    $N_{finished\_worker}$ = num\_finished\_worker() \\
    
    \While {$N_{finished\_worker} > 0$} {
        finished\_worker = get\_finished\_worker() \\
        fitness, individual, current\_steps = finished\_worker.get\_evaluation\_information() \\
        \tcc{Update population with the finished individual}
        population.update(fitness, individual) \\
        $total\_steps \leftarrow total\_steps + current\_steps $\\
        $N_{finished\_worker} \leftarrow N_{finished\_worker} - 1$
    }
}
\end{algorithm}
\vspace{-0.2in}

\begin{algorithm}[h!]
\DontPrintSemicolon
\caption{actor.evaluate($\mathcal{R}$)}
\textbf{Require: } hyperparameter action noise $a_{\text{noise}}$ \\

state = env.reset() \\
done = False \\
steps = 0 \\
total\_reward = 0\\
\While{not \text{done}}{
    action = actor\_network.forward(state) \\
    \If{$a_{\text{noise}} \neq 0$}{
        action = clip(action + $a_{\text{noise}}$ * random.normal(0, 1), -1, 1) \\
    }
    next\_state, reward, done, info = env.step(action) \\
    $\mathcal{R}$.append(state, next\_state, reward, done) \\
    steps $\leftarrow$ steps + 1 \\
    total\_reward $\leftarrow$ total\_reward + 1 \\
    state = next\_state \\
}
return total\_reward, steps
\end{algorithm}

\section{Hyperparameters}
\label{apx:hyper}
Most of hyperparameters are the same value as CEM-RL~\cite{pourchot2018cemrl}. 
However, the size of replay buffer is modified to $2e5$, we analyzed the effect on \adxref{apx:replay}.
Also, we used action noise of $0.1$ as suggested by~\citet{erl2018}. 
\citet{pourchot2018cemrl} reported the action noise is not useful for their algorithm, we found that the action noise improves exploration as the original ERL paper. We use $a_{\text{noise}}=0.1$ for all environments. We discuss the effect on \adxref{apx:actionnoise}.
The $K_{\text{rl}}$ in population control is set to 50.
$p_\text{negative}$ for fixed range algorithms are set to 0. 
We use `1/5' for success rate of $(1+1)$-ES.
Also, we use $p_{\text{desired}}=0.5$ for all environment except Swimmer-v2.
CEM-RL reported that RL algorithms provide deceptive gradients, therefore most of the RL algorithms fail to solve.
Therefore we use $p_{\text{desired}} = 0.1$, which means that the population of the ES is 9 times larger than the RL. 
Consequently, algorithms are become more dependent on the ES part. 

% We use $p_{\text{negative}}=0$ in the fixed range update algorithms, which means that the negative movements are ignored. Further discussion is presented in~\adxref{apx:negative}.
Other values related to the range are presented in~\tabref{tbl:hyperparams}.
We noticed that the corresponding values are about 1/6 of the maximum reward. 
For example, the reward value reaches up to 12000 in HalfCheetah, then the 1/6 of the maximum is about 2000.

\begin{table}[h!]
  \caption{{\bf Hyperparameters} A list of the hyperparameters that vary with the environment  }
  \label{tbl:hyperparams}
%   \vspace{-0.05in}
  \centering
  \begin{tabular}{cc|cc}
    \toprule
    & & \multicolumn{2}{c}{Fixed Range}\\
    \cline{3-4}
    & & Linear & Sigmoid\\
    \hline
    
    \multirow{5}{*}{Range $r$}  & HalfCheetah-v2    & 2000  & 2000 \\
                                & Hopper-v2         & 600   & 600  \\
                                & Walker2D-v2       & 860   & 860  \\
                                & Ant-v2            & 960   & 960  \\
                                & Swimmer-v2        & 48    & 48   \\
                                & Humanoid-v2       & 960   & 960  \\
    \toprule
    & & \multicolumn{2}{c}{Fitness Baseline}\\
    \cline{3-4}
    & & Absolute & Relative\\
    \hline
    \multirow{5}{*}{Baseline $f_b$} & HalfCheetah-v2    & -2000  & 2000 \\
                                    & Hopper-v2         & -600   & 600  \\
                                    & Walker2D-v2       & -860   & 860  \\
                                    & Ant-v2            & -960   & 960  \\
                                    & Swimmer-v2        & -48    & 48   \\
                                    & Humanoid-v2       & -960   & 960  \\
    \bottomrule
  \end{tabular}
%   \vspace{-0.05in}
\end{table}

\newpage

\subsection{Replay Buffer Size}
\label{apx:replay}
The replay buffer size of 200k improves the performance of the proposed algorithm.
The performance of CEM-RL also increased, but not as much as the asynchronous algorithm. 
A possible hypothesis for this phenomenon is as follows. 
In CEM-RL, the critic learning steps are guaranteed in the synchronous stage. 
% Firstly, CEM-RL is a synchronous algorithm, therefore, it guarantees the critic's learning steps, 
However, in AES-RL, the critic is trained in a parallel with the other workers; thus samples should be more productive.
With the reduced size of the replay buffer, it is filled with more recent steps and replace the oldest experiences which is nearly useless. 
Therefore, the samples are more informative.

\tabref{tbl:replay} shows the comparison results. 
In Walker-v2, the size of the replay buffer does not significantly affect the performance of AES-RL. 
However, the performance gap is relatively more significant in HalfCheetah-v2. 
One of the differences of the two environment is that the learning saturates earlier in Walker-v2. 
Therefore, the replay buffer is productive enough. 
Differently, actors in HalfCheetah-v2 still learn at the end of 1M steps. 
Here, We may use other techniques like importance sampling to enhance the efficiency of a mini-batch sample; however, we remain it as future work. 

\begin{table}[h!]
  \caption{{\bf Effect of the Replay Buffer Size} }
%   \vspace{-0.05in}
  \label{tbl:replay}

  \centering
  \begin{tabular}{l|c|cc|cc}
    & $\mu$ & \multicolumn{2}{c|}{Relative Range} & \multicolumn{2}{c}{\multirow{2}{*}{CEM-RL}}\\
    & $\Sigma$ & \multicolumn{2}{c|}{Adaptive} & \multicolumn{2}{c}{}\\
    \cline{2-6}
    & Replay & 200k & 1M & 200k & 1M \\
    \toprule
    \multirow{2}{*}{HalfCheetah-v2} & Mean & \textbf{12550} & 11472 & 11515 & 10725\\
                                    & Std. & \textbf{187}   & 467   & 203   & 354  \\
    \hline
    \multirow{2}{*}{Walker2D-v2}      & Mean & \textbf{5474}  & 5468  & 4503  & 4711   \\
                                    & Std. & \textbf{223}   & 690   & 388   & 155    \\
    
    \bottomrule
  \end{tabular}
%   \vspace{-0.15in}
\end{table}
\subsection{Action Noise}
\label{apx:actionnoise}
CEM-RL reported that the action noise is not useful, as opposed to ERL. 
However it consistently improves the performance a little in our experiments. 
We hypothesize that the $\pi_\mu$ of CEM-RL moves with the weighted average of individuals; therefore, the effect of action noise is reduced. Otherwise, in AES-RL, the action noise increases the chance of better policy which affects directly to the $\pi_\mu$. 

\begin{table}[h!]
  \caption{{\bf Effect of the Action Noise}}
%   \vspace{-0.05in}
  \label{tbl:actionnoise}

  \centering
  \begin{tabular}{l|c|cc}
    & $\mu$ & \multicolumn{2}{c}{Relative Range} \\
    & $\Sigma$ & \multicolumn{2}{c}{Adaptive} \\
    \cline{2-4}
    & $a_{\text{noise}}$& $0.1$ & $0.0$\\
    \toprule
    \multirow{2}{*}{HalfCheetah-v2} & Mean & \textbf{12550} & 12095 \\
                                    & Std. & \textbf{187}   & 338 \\
    \hline
    \multirow{2}{*}{Walker2D-v2}      & Mean & \textbf{5474}  & 5244  \\
                                    & Std. & \textbf{223}   & 670 \\
    
    \bottomrule
  \end{tabular}
%   \vspace{-0.15in/
\end{table}
% \subsection{Affect of the Negative Move}
% \label{apx:negative}

\section{Full Results of Proposed Methods}
\label{apx:result}
We compare all combination of methods proposed in \secref{sec:methods}. 
Except for Swimmer-v2, the relative baseline is the best in both performance and stability. 
In Hopper-v2 and Ant-v2, the relative baseline scores are slightly lower than the best methods, but it is comparable.
Therefore we use the relative baseline methods when compare with the previous algorithms: TD3, CEM, ERL, and CEM-RL. 

In Swimmer-v2, CEM, pure evoluationary algorithm, was the best, ERL, mostly evolutionary algorithm, was also able to solve the environment~\cite{pourchot2018cemrl}. 
Among the asynchronous algorithms, $(1+1)$-ES, which has the most aggressive update rule, is consistently successful. 
From this result we can infer that aggressive exploration is more important in Swimmer-v2.
Also, other methods have a chance to solve the environment, but not always. 

In conclusion, we proposed various update rules for asynchronous algorithms. 
We started from the Rank-Based asynchronous algorithm and the $(1+1)$-ES update algorithm, which are at the extremes. 
Our design purpose is to achieve a balance between the two. 
The relative baseline method with adaptive variance showed the best performance, which effectively balances between aggressive and conservative updates.

\begin{table}[h!]
  \caption{{\bf Full results of our proposed mean-variance update methods} Results are measured with average scores of ten test runs within a total of 1M step from the summation of all worker steps, averaged with ten random seeds. }
  \vspace{-0.05in}
  \label{tbl:fullresult}
  \begin{adjustbox}{max width=\textwidth}
  \centering
  \begin{tabular}{l|c||c|c||c|c|c|c|c|c|c|c}
    \toprule
    \multirow{4}{*}{} & \multirow{2}{*}{Category} & \multicolumn{2}{c||}{Previous algorithms} & \multicolumn{8}{c}{Proposed algorithms} \\
    \cline{3-12}
       &  & (1+1)-ES & Rank-Based  &
    \multicolumn{2}{c|}{Fixed Range} & \multicolumn{2}{c|}{Fitness Baseline} & \multicolumn{2}{c|}{Fixed Range} & \multicolumn{2}{c}{Fitness Baseline} \\
    \cline{2-12}
    & $\mu$ & Full Move & \multirow{2}{*}{Oldest} & Linear & Sigmoid & Absolute & Relative & Linear & Sigmoid & Absolute & Relative \\
    \cline{2-3} \cline{5-12}
    & $\mathbf{\Sigma}$ & Success Rule &    & \multicolumn{4}{c|}{Online Update - Adaptive}& \multicolumn{4}{c}{Online Update - Fixed}  \\
    % \hline
    % \hline
    \midrule
    \multirow{2}{*}{HalfCheetah-v2} & Mean & 11882 & 10010 & 10279 & 12053 & 12224 & \textbf{12550} & 10870 & 12031 & 11767 & 12128  \\
                                    & Std. & 385   & 746   & 1044  & 398   & 422   & \textbf{187}   & 409   & 604   & 458   & 821   \\ 
    \hline
    \multirow{2}{*}{Walker2D-v2}  & Mean & 2347 & 4230 & 5020 & 5360 & 5137 & \textbf{5474} & 3419 & 5121 & 5039 & 5070 \\
                                  & Std. & 320  & 254  & 799  & 683  & 223  & \textbf{223}  & 1676 & 965  & 471  & 557    \\
    \hline
    \multirow{2}{*}{Hopper-v2}  & Mean & 2588 & 3729 & 3506 & 3764 & \textbf{3789} & 3751 & 2996 & 3769 & 3788 & 3423 \\
                                & Std. & 748  & 53   & 179  & 40   & \textbf{26}   & 58   & 1068 & 20   & 30   & 626    \\
    \hline
    \multirow{2}{*}{Ant-v2}     & Mean & 5098 & 3917 & 3015 & \textbf{5140} & 3883 & 5120 & 3406 & 5007 & 4947 & 4613 \\
                                & Std. & 715  & 514  & 1233 & \textbf{546}  & 1047 & 170  & 944  & 891  & 543  & 712    \\
    \hline
    \multirow{2}{*}{Swimmer-v2} & Mean & \textbf{347}  & 59   & 99   & 128  & 97   & 161  & 191  & 81   & 107  & 120  \\
                                & Std. & \textbf{19}   & 8    & 43   & 65   & 32   & 100  & 123  & 12   & 34   & 63   \\
    \hline
    \multirow{2}{*}{Humanoid-v2}& Mean & 600 & 3476 & 5697 & 5958 & 5770 & \textbf{6136} & 5662 & 5695 & 5774 & 5837 \\
                                & Std. & 35  & 1980 & 177  & 301  & 229  & \textbf{444}  & 107  & 209  & 267  & 239  \\
    
    \bottomrule
  \end{tabular}
  \end{adjustbox}
  \vspace{-0.15in}
\end{table}
\section{Ablation Study}
\label{apx:ablation}
AES-RL algorithm mainly consists of three novel methods; an asynchronism, the mean update rule, and the variance update rule. 
In this section, we evaluate the effectiveness of each methods. 

\subsection{Asynchronism}
To compare the effectiveness of asynchronism, we adopt an update rule of CEM-RL based on the simple asynchronous methods in \cite{ANES2013}. 
Therefore, the resulting algorithm is an asynchronous version of CEM-RL, namely ACEM-RL.
As in \cite{ANES2013}, previous results are stored in a separate list with its score.
When a new individual is evaluated, it is stored on the list, and the oldest one is removed to maintain the population size. 
However, there should be a multiplication factor $1/n$ because the update occurs $n$ times frequently. 
Therefore the mean update in \eqnref{eq:covarcalc} is modified to
\begin{equation}
\mu_{t+1} = \frac{1}{n}\sum_{i=1}^{K_e}\lambda_i z_i
\end{equation}

\subsection{Mean and Update Rule}
In the mean update, we already compared various update rules in~\secref{sec:meanvarupdate} and \adxref{apx:result}. 
The result of ACEM-RL, which applies asynchronism to CEM-RL, is in column ``Rank-based"-``Oldest" in \tabref{tbl:fullresult}
In addition, we fix the variance to the constant value. 
Here, We expect that the fixed variance prevents exploration. 

The overall results are displayed in~\tabref{tbl:ablation}. 
As a result, simply applying asynchronism to the previous method without proper mean and variance update reduces the performance.

\begin{table}[h!]
  \caption{{\bf Ablation study} 
  The overall result of the ablation study. 
  From the baseline algorithm CEM-RL, ACEM-RL adopts asynchronism. 
  AES-RL with constant variance means that only the mean is updated. 
  Finally, AES-RL results include all features, asynchronism, mean update, and variance update. }
  \vspace{-0.05in}
  \label{tbl:ablation}

  \centering
  \begin{tabular}{lc|c|c|c|c|c}
    &    & \multirow{3}{*}{CEM-RL}   & \multirow{3}{*}{ACEM-RL}  & \multicolumn{2}{c|}{AES-RL} & \multirow{3}{*}{\textbf{AES-RL}} \\
    &    &                           &                           & \multicolumn{2}{c|}{$\sigma^2=\text{const.}$} & \\
        \cline{5-6}
    &    &                           &                           & 0.0001 & 0.001 & \\
    \toprule
    \multirow{2}{*}{HalfCheetah-v2} & Mean & 10725 & 10010 & 11636 & 12306 & \textbf{12550} \\
                                    & Std. & 397   & 746   & 209   & 233   & \textbf{187} \\
    \hline
    \multirow{2}{*}{Walker2D-v2}    & Mean & 4711  & 4230  & 5167  & 5302  & \textbf{5474} \\
                                    & Std. & 155   & 254   & 251   & 458   & \textbf{223} \\
    
    % & $\mu$ & \multicolumn{3}{c}{Relative Range} \\
    % & $\Sigma$ & Adaptive & $\sigma^2=0.001$  & $\sigma^2=0.0001$\\
    % \toprule
    % \multirow{2}{*}{HalfCheetah-v2} & Mean & \textbf{12550} & 12306 & 11636 \\
    %                                 & Std. & \textbf{187}   & 233   & 209   \\
    % \hline
    % \multirow{2}{*}{Walker-v2}      & Mean & \textbf{5474}  & 5302  & 5167  \\
    %                                 & Std. & \textbf{223}   & 458   & 251   \\
    
    \bottomrule
  \end{tabular}
%   \vspace{-0.15in}
\end{table}

\section{Training Time According to the Number of workers}
\label{apx:numberofworkers}
We compare the execution time for a various number of workers, from 2 to 9. Training time of CEM-RL is measured with original author's implementation. P-CEM-RL is our implementation of parallel version of CEM-RL, which has parallelized actors with synchronous update scheme. For AES-RL, 

\begin{table}[h!]
  \caption{{\bf Training time according to the number of workers} Training time is measured in minutes with Ethernet-connected two machines of Intel i7-6800k with three NVidia GeForce 1080Ti each. }
%   \vspace{-0.1in}
  \label{tbl:numofworker}
\begin{adjustbox}{max width=\textwidth}
  \centering
  \begin{tabular}{lr|c|c|c|c|c|c|c|c|c|c}
    & & CEM-RL & P-CEM-RL & \multicolumn{8}{c}{AES-RL} \\
    \hline
    \multicolumn{2}{r|}{Workers} & 1 & 5 & 2 & 3 & 4 & 5 & 6 & 7 & 8 & 9 \\
    \toprule
    HalfCheetah-v2& & 467.17 & 187.42 & 225.63 & 136.32 & 103.15 & 83.43 & 72.38 & 65.52 & 58.88 & \textbf{54.47} \\
    Walker-v2     & & 487.25 & 205.17 & 275.90  & 163.10  & 131.18 & 105.42 & 97.12  & 84.23 & 81.02 & \textbf{77.33} \\
    Hopper-v2     & & 504.63 & 188.48 & 305.23 & 199.55 & 144.77 & 133.23 & 122.58 & 97.35 & 92.52 & \textbf{90.75} \\

    \bottomrule
  \end{tabular}
  \vspace{-0.15in}
\end{adjustbox}
\end{table}

\section{Contribution of RL and ES in Learning Process}
We approximately compare the contribution of RL and ES agents. 
To measure the contributio,n we used the update ratio $p$ in mean update rules. 
Higher $p$ indicates the new agent moves the mean of the distribution more.
We recorded the $p$ value for all updates, and the values are accumulated for total experiment.
The result in~\tabref{tbl:contribution} shows that the ratio of each agents differs in each environments. 
For Swimmer-v2, our agent fails to find good solution (reward higher than 300) 8 out of 10, we only measured in successful trials. 

% \citet{erl2018} measured similar metric with the ERL algorithm. 

% \begin{table}[h!]
%   \caption{{\bf Contribution of RL and ES\KH{Modify}}}
% %   \vspace{-0.05in}
%   \label{tbl:contribution}

%   \centering
%   \begin{tabular}{c|cc}
%   & ES (\%) & RL (\%) \\
%   \toprule
%   HalfCheetah-v2 &        &         \\
%   Walker2D-v2    &        &         \\
%   Hopper-v2      &        &         \\
%   Ant-v2         &        &         \\
%   Swimmer-v2     &        &         \\
%   Humanoid-v2    &        &         \\
%   \bottomrule
%   \end{tabular}
% %   \vspace{-0.15in}
% \end{table}

\begin{table}[h!]
  \caption{{\bf Contribution of RL and ES}}
  \vspace{-0.05in}
  \label{tbl:contribution}
  \begin{adjustbox}{max width=\textwidth}
  \centering
  \begin{tabular}{l|cccccc}
   & HalfCheetah-v2 & Walker2D-v2 & Hopper-v2 & Ant-v2 & Swimmer-v2 & Humanoid-v2 \\
   \toprule
   ES (\%) & 37.1 & 50.3 & 64.2 & 22.5 & 79.5 & 50.4 \\
   RL (\%) & 62.9 & 49.7 & 35.8 & 77.5 & 20.5 & 49.6 \\ 
   \bottomrule
  \end{tabular}
  \vspace{-0.15in}
  \end{adjustbox}
\end{table}

\section{Step-wise Learning Curve}
\label{adx:steplearningcurve}
\begin{figure}[h!]
\centering
\begin{adjustbox}{max width=\textwidth}
\begin{tabular}{ccc}
    % \vspace{-0.2in}
    \includegraphics{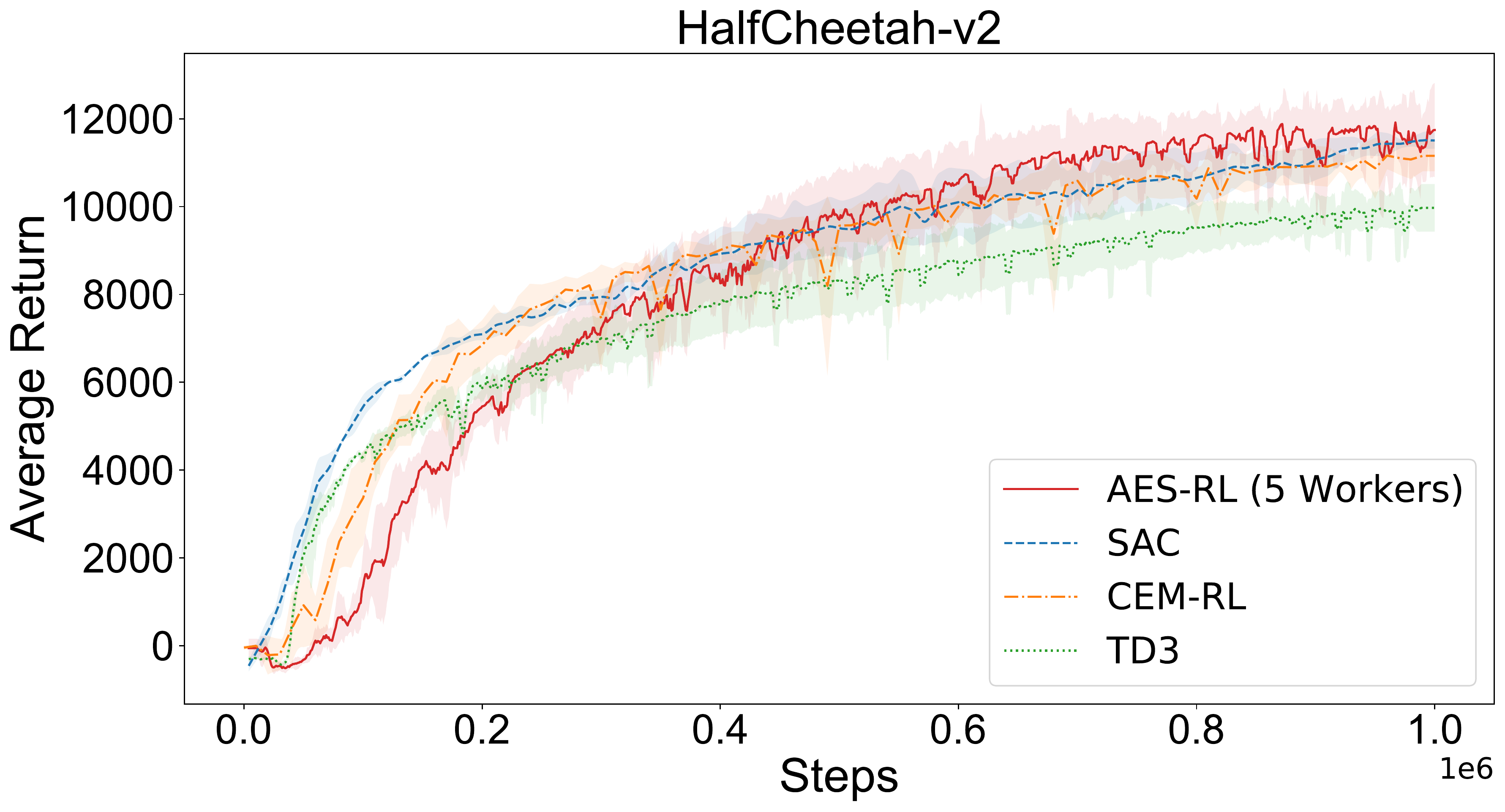} &
    \includegraphics{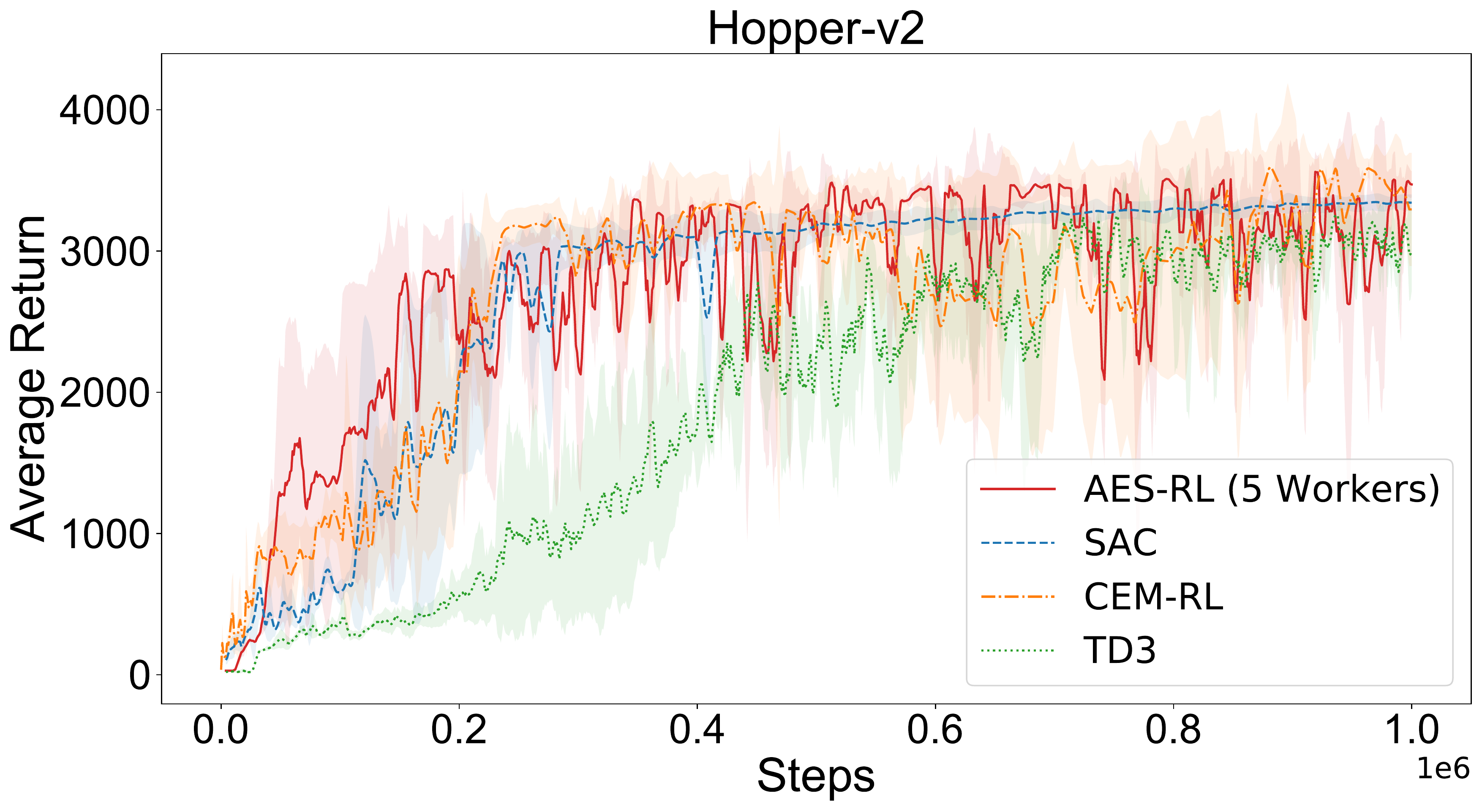} &
    \includegraphics{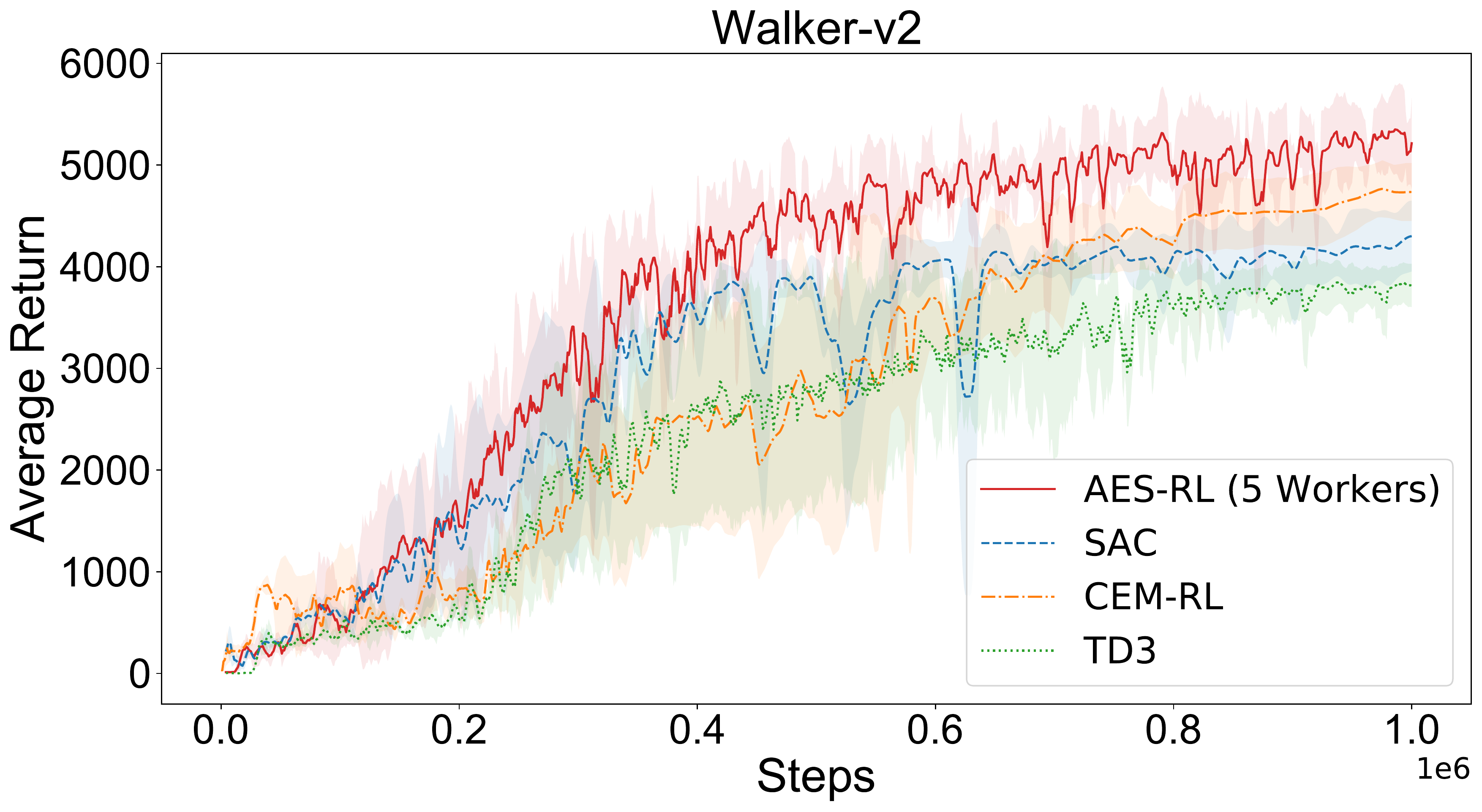} \\
    \includegraphics{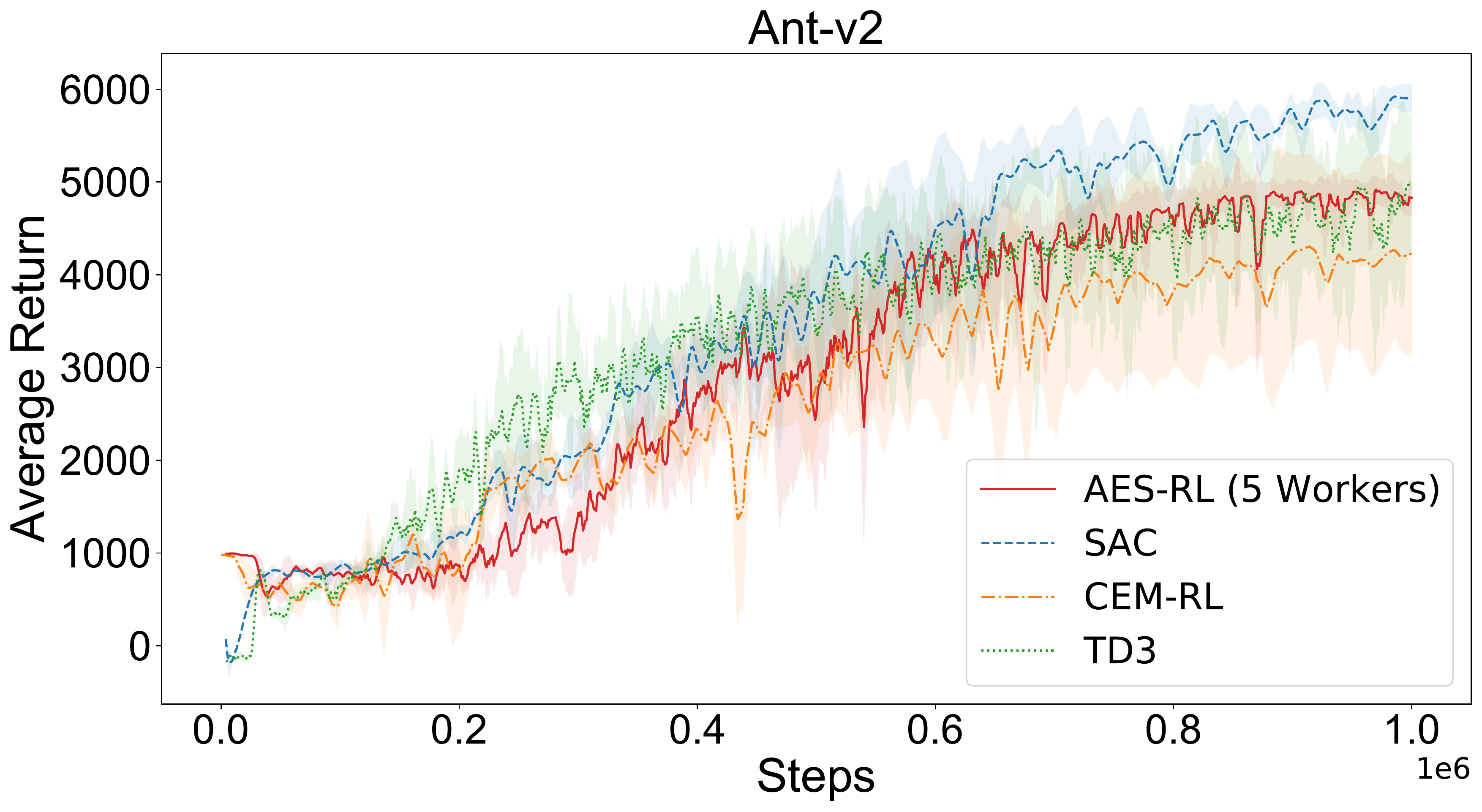} &
    \includegraphics{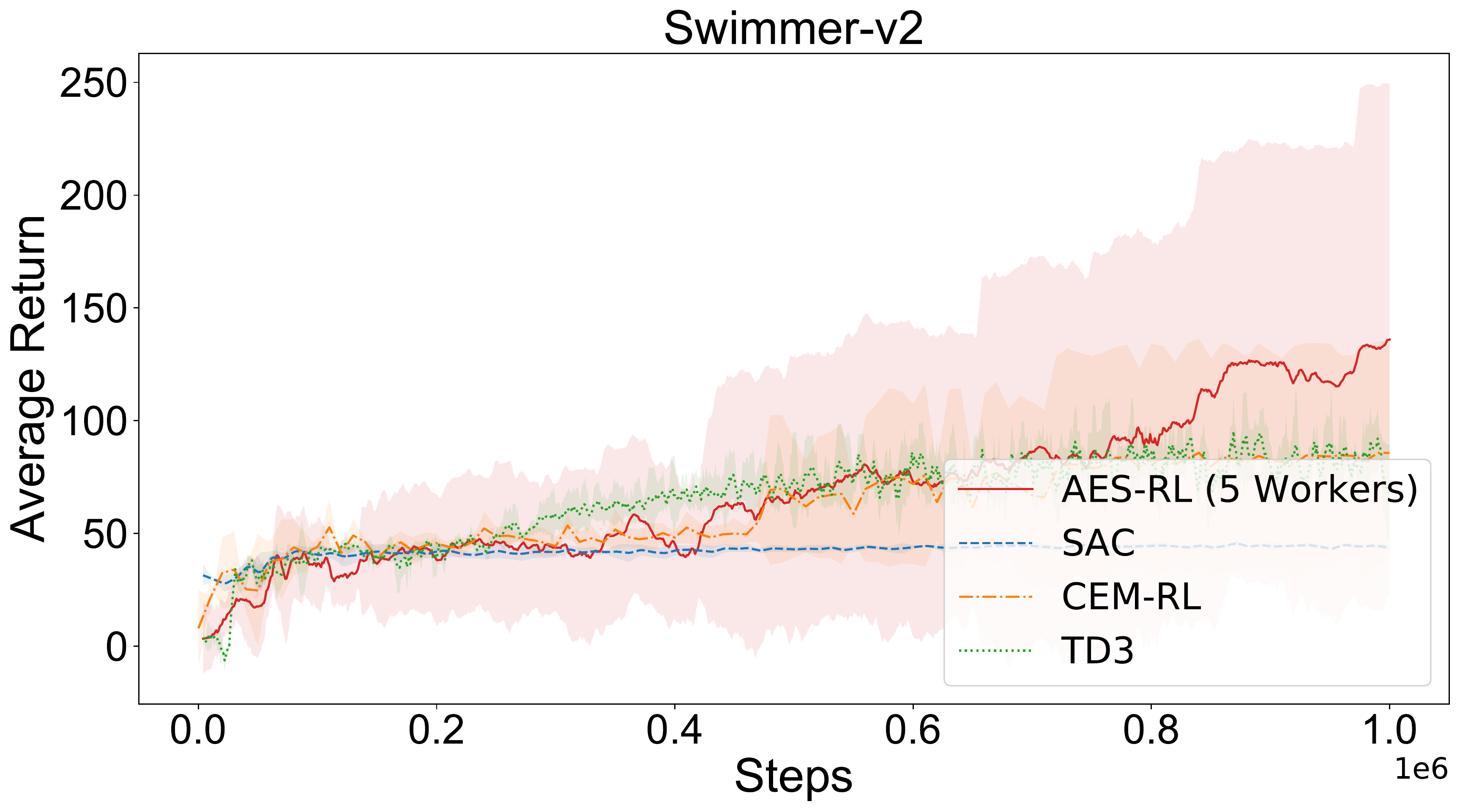} &
    \includegraphics{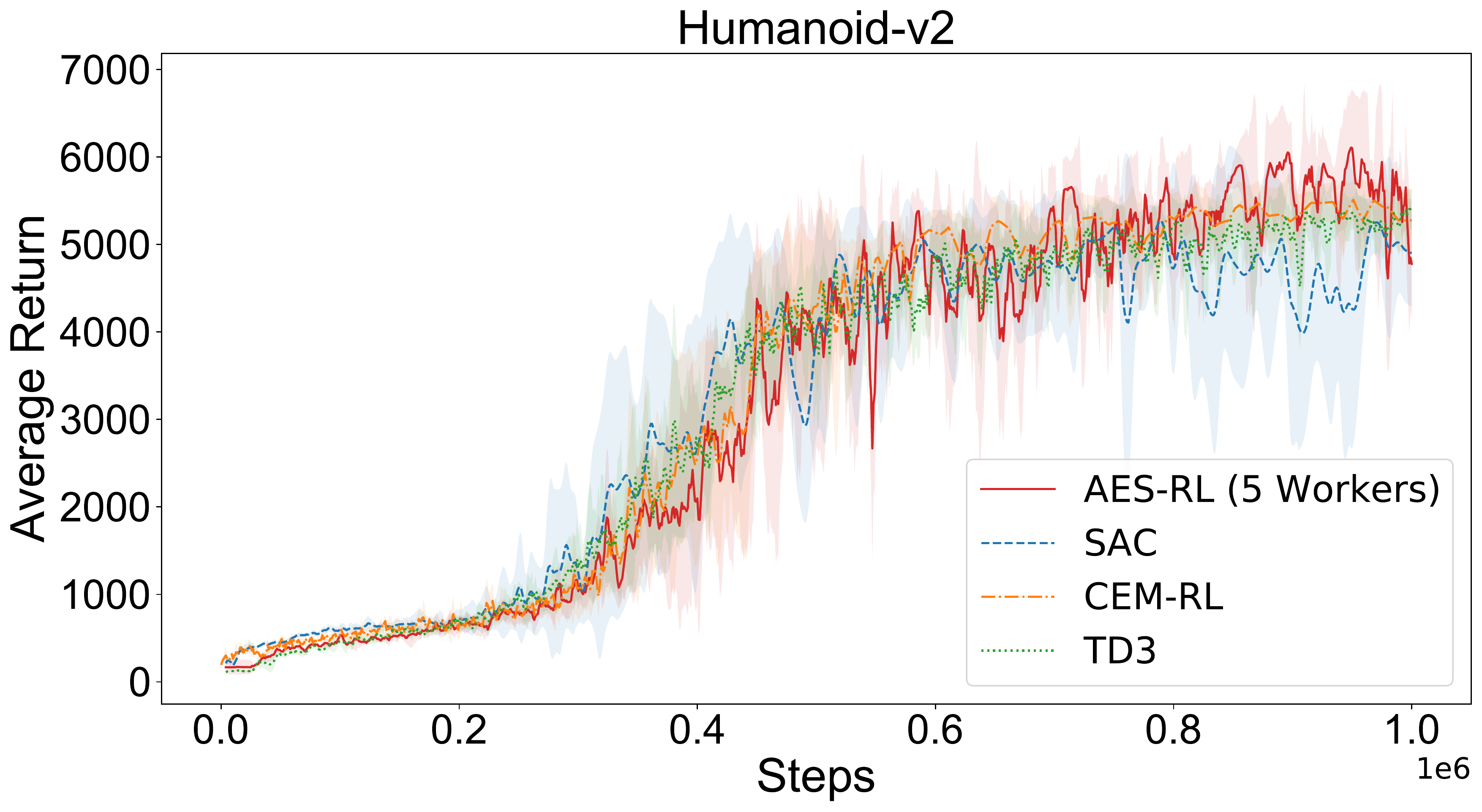} \\
    \vspace{-0.2in}
\end{tabular} 
\end{adjustbox}
\end{figure}

\end{document}